%% file: main.tex
\documentclass{styles/svproc}
\usepackage{url}
\usepackage{amssymb}
\usepackage[cmex10]{amsmath}

\usepackage{subfigure}
\usepackage{epsfig}
\usepackage[ruled]{algorithm2e}
\usepackage{algorithmic}

\usepackage{mathtools}
\DeclarePairedDelimiter{\ceil}{\lceil}{\rceil}
\usepackage{wrapfig}

\usepackage{color}

\newcommand{\rev}[1]{{\color{black}{#1}}}
\renewcommand{\max}{\text{max}}
\renewcommand{\min}{\text{min}}

\newcommand{\goodgap}{%
\hspace{\subfigtopskip}%
\hspace{\subfigbottomskip}}

\begin{document}
\mainmatter              

\title{An Approximation Algorithm for Risk-averse Submodular Optimization}
\titlerunning{CVaR Submodular Maximization}  
%
\author{Lifeng Zhou \and Pratap Tokekar}
%
\authorrunning{L. Zhou \&  P. Tokekar}
%
%
\institute{Department of Electrical \& Computer Engineering, Virginia Tech, USA.\\
\email{\{lfzhou, tokekar\}@vt.edu}}

\maketitle              

\begin{abstract}
We study the problem of incorporating risk while making combinatorial decisions under uncertainty. We formulate a discrete submodular maximization problem for selecting a set using Conditional-Value-at-Risk (CVaR), a risk metric commonly used in financial analysis. While CVaR has recently been used in optimization of linear cost functions in robotics, we take the first stages towards extending this to discrete submodular optimization and provide several positive results. Specifically, we propose the Sequential Greedy Algorithm that provides an approximation guarantee on finding the maxima of the CVaR cost function under a matroidal constraint. The approximation guarantee shows that the solution produced by our algorithm is within a constant factor of the optimal and an additive term that depends on the optimal. Our analysis uses the curvature of the submodular set function, and proves that the algorithm runs in polynomial time. This formulates a number of combinatorial optimization problems that appear in robotics. We use two such problems, vehicle assignment under uncertainty for mobility-on-demand and sensor selection with failures for environmental monitoring, as case studies to demonstrate the efficacy of our formulation. 
\end{abstract}
\section{Introduction}\label{sec:intro}
Combinatorial optimization problems find a variety of applications in robotics. Typical examples include:
\begin{itemize}
\item \emph{Sensor placement:} Where to place sensors to maximally cover the environment~\cite{o1987art} or reduce the uncertainty in the environment~\cite{krause2008near}?   
\item \emph{Task allocation:} How to allocate tasks to robots to maximize the overall utility gained by the robots~\cite{gerkey2004formal}?
\item \emph{Combinatorial auction:} How to choose a combination of items for each player to maximize the total rewards~\cite{vondrak2008optimal}?
\end{itemize}
Algorithms for solving such problems find use in sensor placement for environment monitoring~\cite{o1987art,krause2008near}, robot-target assignment and tracking~\cite{spletzer2003dynamic,tekdas2010sensor,tokekar2014multi}, and informative path planning~\cite{singh2009efficient}. The underlying optimization problem in most cases can be written as:
\begin{equation}
\underset{{\mathcal{S} \in \mathcal{I}, \mathcal{S}\in \mathcal{X}}}{\text{max}} f(\mathcal{S}),
\label{eqn:basic}
\end{equation}
where $\mathcal{X}$ denotes a ground set from which a subset of elements $S$ must be chosen. $f$ is a monotone submodular utility function~\cite{nemhauser1978analysis,fisher1978analysis}. Submodularity is the property of diminishing returns. Many information theoretic measures, such as mutual information~\cite{krause2008near}, and geometric measures such as the visible area~\cite{ding2017multi}, are known to be submodular. $\mathcal{I}$ denotes a matroidal constraint~\cite{nemhauser1978analysis,fisher1978analysis}. Matroids are a powerful combinatorial tool that can represent constraints on the solution set, e.g., cardinality constraints (``place no more than $k$ sensors'') and connectivity constraints (``the communication graph of the robots must be connected'')~\cite{williams2017decentralized}. The objective of this problem is to find a set $\mathcal{S}$ satisfying a matroidal constraint $\mathcal{I}$ and maximizing the utility $f(\mathcal{S})$. The general form of this problem is NP-complete. However, a greedy algorithm yields a constant factor approximation guarantee~\cite{nemhauser1978analysis,fisher1978analysis}.

In practice, sensors can fail or get compromised~\cite{wood2002denial} or robots may not know the exact positions of the targets~\cite{dames2017detecting}. Hence, the utility $f(\mathcal{S})$ is not necessarily deterministic but can have uncertainty. Our main contribution is to extend the traditional formulation given in Eq.~\ref{eqn:basic} to also account for the uncertainty in the actual cost function. We model the uncertainty by assuming that the utility function is of the form $f(\mathcal{S},y)$ where $\mathcal{S}\in \mathcal{X}$ is the decision variable and $y\in \mathcal{Y}$ represents a random variable \rev{which is independent of $\mathcal{S}$.} We focus on the case where $f(\mathcal{S},y)$ is monotone submodular in $\mathcal{S}\in \mathcal{X}$ and integrable in $y$.

The traditional way of stochastic optimization is to use the expected utility as the objective function: 
${\max}_{\mathcal{S} \in \mathcal{I}, \mathcal{S}\in \mathcal{X}} {\mathbb{E}_y} [f(\mathcal{S},y)].$
Since the sum of the monotone submodular functions is  monotone submodular, $\mathbb{E}_y [f(\mathcal{S},y)]$ is still monotone submodular in $\mathcal{S}$. Thus, the greedy algorithm still retains its constant-factor performance guarantee~\cite{nemhauser1978analysis,fisher1978analysis}. Examples of this approach include  influence maximization~\cite{kempe2003maximizing}, moving target detection and tracking~\cite{dames2017detecting}, and robot assignment with travel-time uncertainty~\cite{prorok2018supermodular}. 

\begin{wrapfigure}{r}{0.5\textwidth}
  \begin{center}
    \includegraphics[width=0.5\textwidth]{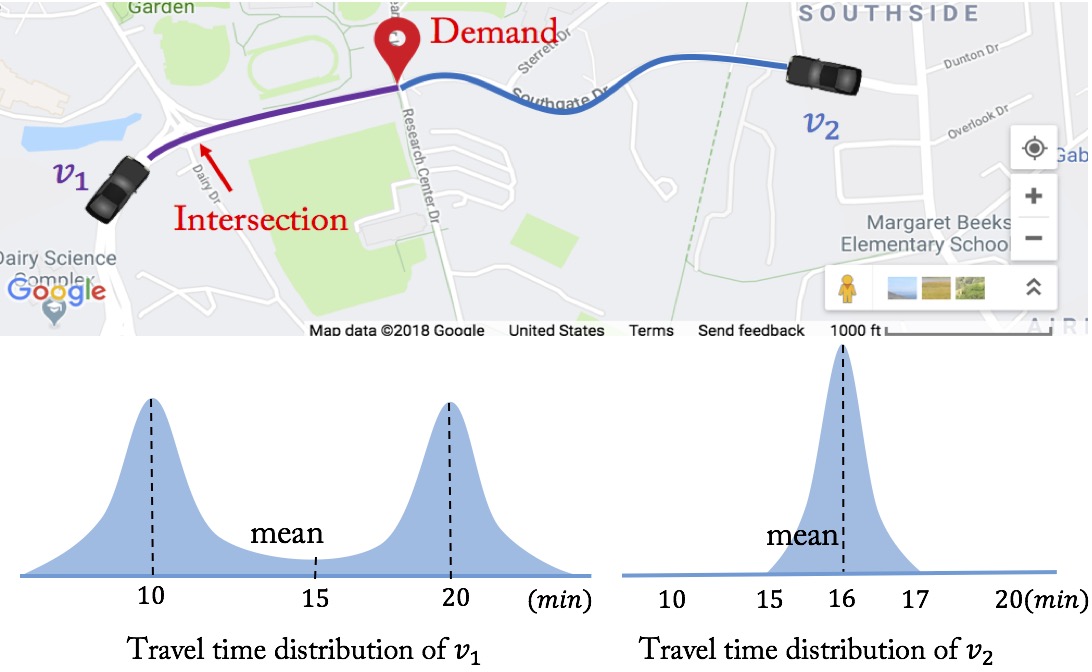}
  \end{center}
  \caption{Mobility on demand with travel time uncertainty of self-driving vehicles.\label{fig:simple_mod}}
\end{wrapfigure}

While optimizing the expected utility has its uses, it also has its pitfalls. Consider the example of mobility-on-demand where two self-driving vehicles, $v_1$ and $v_2$, are available to pick up the passengers at a demand location (Fig.~\ref{fig:simple_mod}).  $v_1$ is closer to the demand location, but it needs to cross an intersection where it may need to stop and wait. $v_2$ is further from the demand location but there is no intersection along the path. The travel time for $v_1$ follows a bimodal distribution (with and without traffic stop) whereas that for $v_2$ follows a unimodal distribution with a higher mean but lower uncertainty. Clearly, if the passenger uses the expected travel time as the objective, they would choose $v_1$. However, they will risk waiting a much longer time, i.e., $17 \sim 20min$ about half of the times. A more risk-averse passenger would choose $v_2$ which has higher expected waiting time $16min$ but a lesser risk of waiting longer.

Thus, in these scenarios, it is natural to go beyond expectation  and focus on a risk-averse measure. One popular coherent risk measure is \emph{Conditional-Value-at-Risk} (CVaR)~\cite{pflug2000some,rockafellar2000optimization}. CVaR takes a risk level $\alpha$ which is the probability of the worst $\alpha$-tail cases. Loosely speaking, maximizing CVaR is equivalent to maximizing the expectation of the worst $\alpha$-tail scenarios.\footnote{We formally review CVaR and other related concepts in Section~\ref{subsec:risk}} This risk-averse decision is rational especially when the failures can lead to unrecoverable consequences, such as a sensor failure.


\textbf{Related work}. Yang and Chakraborty studied a chance-constrained combinatorial optimization problem that takes into account the risk in multi-robot assignment~\cite{yang2017algorithm}. They later extended this to knapsack problems~\cite{yang2018algorithm}. They solved the problem by transforming it to a risk-averse problem with mean-variance measure~\cite{markowitz1952portfolio}. Chance-constrained optimization is similar to optimizing the Value-at-Risk (VaR), which is another popular risk measure in finance~\cite{morgan1996riskmetrics}. However, Majumdar and Pavone argued that CVaR is a better measure to quantify risk than VaR or mean-variance based on six proposed axioms in the context of robotics~\cite{majumdar2017should}.  

Several works have focused on optimizing CVaR. In their seminal work~\cite{rockafellar2000optimization}, Rockafellar and Uryasev presented an algorithm for CVaR minimization for reducing the risk in financial \emph{portfolio optimization} with a large number of instruments. Note that, in {portfolio optimization}, we select a distribution over available decision variables, instead of selecting a single one. Later, they showed the advantage of optimizing CVaR for general loss distributions in finance~\cite{rockafellar2002conditional}. 

When the utility is a discrete submodular set function, i.e., $f(\mathcal{S},y)$, 
Maehara presented a negative result for maximizing $\text{CVaR}$~\cite{maehara2015risk}---  there is no polynomial time multiplicative approximation algorithm for this problem under some reasonable assumptions in computational complexity. To avoid this difficulty, Ohsaka and Yoshida in~\cite{ohsaka2017portfolio} used the same idea from {portfolio optimization} and proposed a method of selecting a distribution over available sets rather than selecting a single set, and gave a provable guarantee. Following this line, Wilder considered a CVaR maximization of a {continuous submodular} function instead of the submodular set functions~\cite{wilder2018risk}. They gave a $(1 - 1/e)$--approximation algorithm for {continuous submodular} functions. They also evaluated the algorithm for discrete submodular  functions using {portfolio optimization}~\cite{ohsaka2017portfolio}.

\textbf{Contributions}. We focus on the problem of selecting a single set, similar to~\cite{maehara2015risk}, to maximize CVaR rather than portfolio optimization~\cite{ohsaka2017portfolio,wilder2018risk}. This is because we are motivated by applications where a one-shot decision (placing sensors and assigning vehicles) must be taken. Our contributions are as follows: 
\begin{itemize}
\item We propose the Sequential Greedy Algorithm (SGA) which uses the deterministic greedy algorithm~\cite{nemhauser1978analysis,fisher1978analysis} as a subroutine to find the maximum value of CVaR (Algorithm~\ref{alg:sga}).  
\item We prove that the solution found by SGA is within a constant factor of the optimal performance along with an additive term which depends on the optimal value. We also prove that SGA runs in polynomial time (Theorem~\ref{thm:appro_bound_compu}) and the performance improves as the running time increases.
\item We demonstrate the utility of the proposed CVaR maximization problem through two case studies (Section~\ref{subsec:case_study}). We evaluate the performance of SGA through simulations (Section~\ref{sec:simulation}). 
\end{itemize}

\textbf{Organization of rest of the paper}. We give the necessary background knowledge for the rest of the paper in Section~\ref{sec:background}. We formulate   the CVaR submodular maximization problem with two case studies in Section~\ref{sec:problem_case}. We present SGA along with the analysis of its computational complexity and approximation ratio in Section~\ref{sec:alg_ana}. We illustrate the performance of  SGA to the two case studies in Section~\ref{sec:simulation}. We conclude the paper in Section~\ref{sec:conclue}.

\section{Background and Preliminaries}\label{sec:background}
\begin{wrapfigure}{r}{0.5\textwidth}
  \begin{center}
    \includegraphics[width=0.5\textwidth]{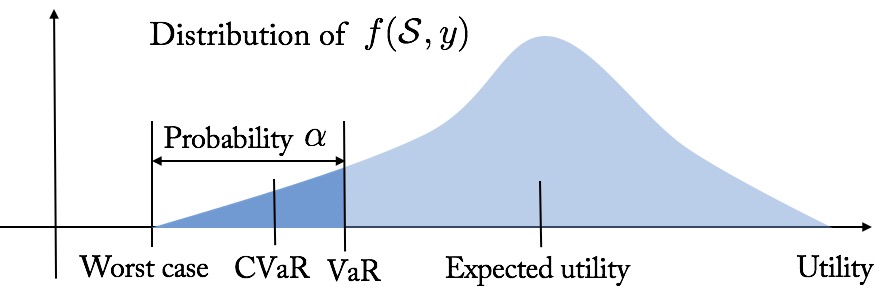}
  \end{center}
  \caption{An illustration of risk measures: VaR and CVaR.\label{fig:var_cvar}}
\end{wrapfigure}
We start by defining the conventions used in the paper. 

Calligraphic font denotes a set (e.g., $\mathcal{A}$).  Given a set $\mathcal{A}$,  $2^{\mathcal{A}}$ denotes its power set. $|\mathcal{A}|$ denotes the cardinality of $\mathcal{A}$. Given a set $\mathcal{B}$, $\mathcal{A}\setminus\mathcal{B}$ denotes the set of elements in $\mathcal{A}$ that are not in~$\mathcal{B}$. $\text{Pr}[\cdot]$ denotes the probability of an event and $\mathbb{E}[\cdot]$ denotes the expectation of a random variable.  $\ceil{x} = \min\{n\in\mathbb{Z}|x\leq n\}$ where  $\mathbb{Z}$ denotes the set of integer. 

Next, we give the background on set functions (in the appendix file) and risk measures. 
\subsection{Risk measures}\label{subsec:risk}
Let $f(\mathcal{S}, y)$ be a utility function with decision set $\mathcal{S}$ and the random variable $y$. For each $\mathcal{S}$, the utility $f(\mathcal{S}, y)$ is also a random variable with a distribution induced by that of $y$. First, we define the Value-at-Risk at risk level $\alpha \in (0, 1]$.

\noindent\textbf{Value at Risk:}
\begin{equation}
\text{VaR}_{\alpha}(\mathcal{S}) = \text{inf} \{\tau\in\mathbb{R}, \text{Pr} [f(\mathcal{S},y)\leq \tau] \geq \alpha\}.
\label{eqn:VaR}
\end{equation}
Thus, $\text{VaR}_{\alpha}(\mathcal{S})$ denotes the left endpoint of the $\alpha$-quantile(s) of the random variable $f(\mathcal{S},y)$. The Conditional-Value-at-Risk is the expectation of this set of $\alpha$-worst cases of $f(\mathcal{S}, y)$, defined as:

\noindent\textbf{Conditional Value at Risk:}
\begin{equation}
\text{CVaR}_{\alpha}(\mathcal{S}) = \underset{y}{\mathbb{E}}[f(\mathcal{S},y)|f(\mathcal{S},y)\leq \text{VaR}_{\alpha}(\mathcal{S})].
\label{eqn:CVaR}
\end{equation} 
Fig.~\ref{fig:var_cvar} shows an illustration of $\text{VaR}_{\alpha}(\mathcal{S})$ and $\text{CVaR}_{\alpha}(\mathcal{S})$. $\text{CVaR}_{\alpha}(\mathcal{S})$ is more popular than $\text{VaR}_{\alpha}(\mathcal{S})$ since it has better properties~\cite{rockafellar2000optimization}, such as  \emph{coherence}~\cite{artzner1999coherent}. 

When optimizing $\text{CVaR}_{\alpha}(\mathcal{S})$, we usually resort to an auxiliary function:
 $$H(\mathcal{S}, \tau) = \tau - \frac{1}{\alpha}\mathbb{E}[(\tau-f(\mathcal{S},y))_{+}].$$
We know that optimizing $\text{CVaR}_{\alpha}(\mathcal{S})$ over $\mathcal{S}$ is equivalent to optimizing the auxiliary function $H(\mathcal{S}, \tau)$ over $\mathcal{S}$ and $\tau$~\cite{rockafellar2000optimization}. The following lemmas give useful properties of the auxiliary function $H(\mathcal{S}, \tau)$.

\begin{lemma}
If $f(\mathcal{S},y)$ is normalized, monotone increasing and submodular in set $\mathcal{S}$ for any realization of $y$, the auxiliary function $H(\mathcal{S},\tau)$ is monotone increasing and submodular, but not necessarily normalized in set $\mathcal{S}$ for any given $\tau$.
\label{lem:auxiliary_function}
\end{lemma}
We provide the proofs for all the Lemmas and Theorem in the appendix file.
\begin{lemma}
The auxiliary function $H(\mathcal{S},\tau)$ is concave in $\tau$ for any given set $\mathcal{S}$. 
\label{lem:auxi_concave}
\end{lemma}
\begin{lemma}
For any given set $\mathcal{S}$, the gradient of the  auxiliary function $H(\mathcal{S},\tau)$ with respect to $\tau$ fulfills: $-(\frac{1}{\alpha}-1) \leq \frac{\partial H(\mathcal{S},\tau)}{\partial \tau} \leq 1$. 
\label{lem: gradient_auxiliary_function}
\end{lemma}
\section{Problem Formulation and Case Studies} \label{sec:problem_case}
We first formulate the CVaR submodular maximization problem and then present two applications which we use as case studies. 
\subsection{Problem Formulation}\label{subsec:problem}
\noindent\textbf{CVaR Submodular Maximization}: We consider the problem of maximizing $\text{CVaR}_{\alpha}(\mathcal{S})$ over a decision set $\mathcal{S}\subseteq \mathcal{X}$ under a matroid constraint $\mathcal{S}\in \mathcal{I}$.  We know that maximizing $\text{CVaR}_{\alpha}(\mathcal{S})$ over $\mathcal{S}$ is equivalent to maximizing the auxiliary function $H(\mathcal{S}, \tau)$ over $\mathcal{S}$ and $\tau$~\cite{rockafellar2000optimization}. Thus, we propose the maximization problem as:

\begin{problem}
\begin{eqnarray}
\max ~~\tau - \frac{1}{\alpha}\mathbb{E}[(\tau-f(\mathcal{S},y))_{+}]\nonumber\\
s.t.~~\mathcal{S} \in \mathcal{I}, \mathcal{S}\subseteq \mathcal{X}, \tau\in[0, \Gamma],
\label{eqn:cvar_max}
\end{eqnarray}
\label{pro:cvar_max}
\end{problem}
\rev{where $\Gamma$ is the upper bound of the parameter $\tau$.}
Problem~\ref{pro:cvar_max} gives a risk-averse version of maximizing submodular set functions. 

\subsection{Case Studies}\label{subsec:case_study}
The risk-averse submodular maximization has many applications, as it has been written in Section~\ref{subsec:case_study}. We describe two specific applications which we will use in the simulations. 

\subsubsection{Resilient Mobility-on-Demand}\label{subsubsec:mod}

\begin{wrapfigure}{r}{0.5\textwidth}
  \begin{center}
    \includegraphics[width=0.5\textwidth]{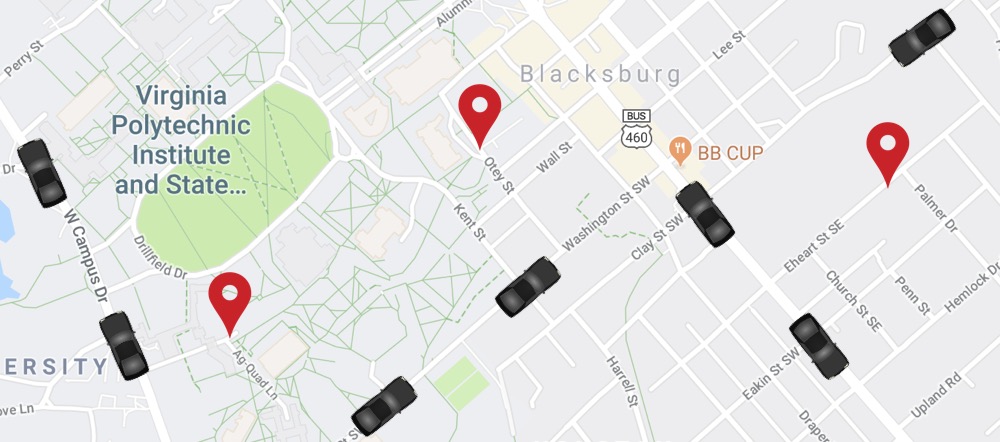}
  \end{center}
  \caption{Mobility-on-demand with multiple demands and multiple self-driving vehicles. \label{fig:MoD}}
\end{wrapfigure}

Consider a mobility-on-demand problem where we assign $R$ vehicles to $N$ demand locations under arrival-time uncertainty. An example is shown in Fig.~\ref{fig:MoD} where seven self-driving vehicles must be assigned to three demand locations to pick up passengers. We follow the same constraint setting in~\cite{prorok2018supermodular}--- each vehicle can be assigned to at most one demand but multiple vehicles can be assigned to the same demand. Only the vehicle that arrives first is chosen for picking up the passengers. Note that the advantage of the redundant assignment to each demand is that it counters the effect of uncertainty and reduces the waiting time at demand locations~\cite{prorok2018supermodular}.  \rev{This may be too conservative for consumer mobility-on-demand services but can be crucial for urgent and time-critical tasks such as delivery medical supplies~\cite{ackerman2018medical}.}

Assume the arrival time for the robot to arrive at demand location is a random variable. The distribution can depend on the mean-arrival time. For example, it is possible to have a shorter path that passes through many intersections, which \rev{leads to} an uncertainty on arrival time. While a longer road (possibly a highway) has a lower arrival time uncertainty. Note that for each demand location, there is a set of robots assigned to it. The vehicle selected at the demand location is the one that arrives first. 
Then, this problem becomes a minimization one since we would like to minimize the  arrival time at all demand locations. We convert it into a maximization one by taking the reciprocal of the arrival time. Specifically, we use the arrival \emph{efficiency} which is the reciprocal of arrival time. Instead of selecting the vehicle at the demand location with minimum arrival time, we select the vehicle with maximum arrival efficiency. The arrival efficiency is also a random variable, and has a distribution depending on mean-arrival efficiency. Denote the arrival efficiency for robot $j\in\{1,...,R\}$ arriving at demand location $i\in\{1,...,N\}$ as $e_{ij}$. Denote the assignment utility as the arrival efficiency at all locations, that is, 
\begin{equation}
f(\mathcal{S}, y)  = \sum_{i\in N} \text{max}_{j\in \mathcal{S}_i}e_{ij}
\label{eqn:fsy_assign}
\end{equation} with $\bigcup_{i=1}^{N} \mathcal{S}_i = \mathcal{S}$ and $\mathcal{S}_i \cap \mathcal{S}_k =  \emptyset, ~i, k \in \{1, \cdots, N\}$. $\mathcal{S}_i \cap \mathcal{S}_k =  \emptyset$ indicates the selected set $\mathcal{S}$ satisfies a partition matroid constraint, $\mathcal{S} \in \mathcal{I}$, which represents that each robot can be assigned to at most one demand.  The assignment utility $f(\mathcal{S}, y)$ is monotone submodular in $\mathcal{S}$ due to the ``max'' function. $f(\mathcal{S}, y)$ is normalized since $f(\emptyset, y)=0$. Here, we regard the uncertainty as a risk. Our risk-averse assignment problem is a trade-off between efficiency and uncertainty. Our goal is to maximize the total efficiencies at the demand locations while considering the risk from uncertainty. 
\subsubsection{Robust Environment Monitoring} \label{subsubsec:environment_monitoring}
\begin{figure*}[htb]
\centering{
\subfigure[Part of Virgina Tech campus from Google Earth.]{\includegraphics[width=0.48\columnwidth]{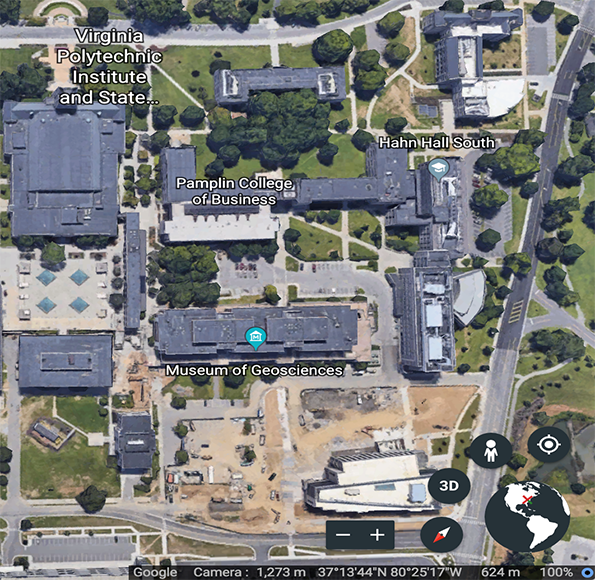}}\goodgap
\subfigure[Top view of part of a campus and ground sensor's visibility region.]{\includegraphics[width=0.48\columnwidth]{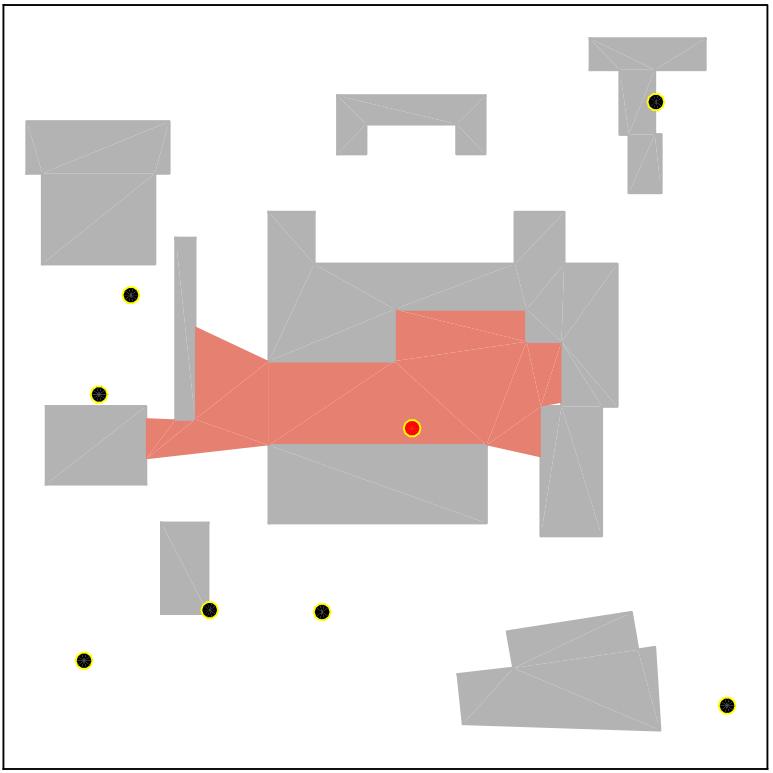}}
\caption{ Campus monitoring by using a set of sensors with visibility regions.
\label{fig:environment_monitor}}}
\end{figure*}

Consider an environment monitoring problem where we monitor part of a campus with a group of ground sensors (Fig.~\ref{fig:environment_monitor}). Given a set of $N$ candidate positions $\mathcal{X}$, we would like to choose a subset of $M$ positions $ \mathcal{S}\subseteq \mathcal{X}, M\leq N$, to place visibility-based sensors to maximally cover the environment. The visibility regions of the ground sensors are obstructed by the buildings in the environment (Fig.~\ref{fig:environment_monitor}-(b)). Consider a scenario where the probability of failure of a sensor depends on the area it can cover. That is, a sensor covering a larger area has a larger risk of failure associated with it. This may be due to the fact that the same number of pixels are used to cover a larger area and therefore, each pixel covers proportionally a smaller footprint. As a result, the sensor risks missing out on detecting small objects. 

Denote the probability of success and the visibility region for each sensor $i, i \in\{1,...,N\}$ as $p_i$ and $v_i$, respectively. Thus, the polygon each sensor $i$  monitors is also a random variable. Denote this random polygon as $A_i$ and denote the selection utility as the joint coverage area of a set of sensors, $\mathcal{S}$, that is,
\begin{equation}
f(\mathcal{S}, y)  = \text{area}(\bigcup_{i=1:M} A_i), ~i\in\mathcal{S}, \mathcal{S} \subseteq \mathcal{I}. 
\label{eqn:fsy_covering}
\end{equation}
The selection utility $f(\mathcal{S}, y)$ is monotone submodular in $\mathcal{S}$ due to the overlapping area. $f(\mathcal{S}, y)$ is normalized since $f(\emptyset, y)=0$. Here, we regard the sensor failure as a risk. Our robust environment monitoring problem is a trade-off between area coverage and sensor failure. Our goal is to maximize the joint-area covered while considering the risk from sensor failure.  

\section{Algorithm and Analysis}\label{sec:alg_ana}

We present the Sequential Greedy Algorithm (SGA) for solving Problem~\ref{pro:cvar_max} by leveraging the useful properties of the auxiliary function $H(\mathcal{S}, \tau)$. The pseudo-code  is given in Algorithm~\ref{alg:sga}. SGA mainly consists of searching for the appropriate value of $\tau$ by solving a subproblem for a fixed $\tau$ under a matroid constraint. Even for a fixed $\tau$, the subproblem of optimizing the auxiliary function is NP-complete. Nevertheless, we can employ the greedy algorithm for the subproblem, and sequentially apply it for searching over all $\tau$. We explain each stage in detail next.
\subsection{Sequential Greedy Algorithm}
These are four stages in SGA: 
\paragraph{a) Initialization (line~\ref{line:initiliaze}):} Algorithm~\ref{alg:sga} defines a storage set $\mathcal{M}$ and initializes it to be the empty set. Note that, for each specific $\tau$, we can use the greedy algorithm to obtain a near-optimal solution $\mathcal{S}^{G}$ based on the monotonicity and submodularity of the auxiliary function $H(\mathcal{S}, \tau)$. $\mathcal{M}$ stores all the $(\mathcal{S}^{G}, \tau)$ pairs when searching all the possible values of $\tau$. 

\paragraph{b) Searching for $\tau$ (\textbf{for} loop in lines~\ref{line:search_tau_forstart}--\ref{line:search_tau_forend}):} We use a user-defined separation $\Delta$ (line~\ref{line:search_tau_separation}) to sequentially search for all possible values of $\tau$ within $[0, \Gamma]$. 
$\Gamma$ is an upper bound on $\tau$ and can be set by the user based on the specific problem at hand. We show how to find $\Gamma$ for the specific cases in Section~\ref{sec:simulation}. 

\paragraph{c) Greedy algorithm (lines~\ref{line:gre_empty}--\ref{line:gre_while_end}):} For a specific $\tau$, say $\tau_i$, we use the greedy approach to choose  set $\mathcal{S}_{i}^{G}$. We first initialize set $\mathcal{S}_{i}^{G}$ to be the empty set (line~\ref{line:gre_empty}). Under a matroid constraint, $\mathcal{S}_{i}^{G} \in \mathcal{I}$ (line~\ref{line:gre_while_start}), we add a new element $s$ which gives the maximum marginal gain of $H(\mathcal{S}_{i}^{G}, \tau_i)$ (line~\ref{line:gre_while_margin}) into set $\mathcal{S}_{i}^{G}$ (line~\ref{line:gre_while_pick}) in each round.

\paragraph{d) Find the best pair (line~\ref{line:find_best_pair}):} Based on the collection of pairs $\mathcal{M}$ (line~\ref{line:pair_collection}), we pick the pair $(\mathcal{S}_{i}^{G}, \tau_i) \in \mathcal{M}$ that maximizes $H(\mathcal{S}_{i}^{G}, \tau_i)$ as the final solution ${S}_{i}^{G}$. We denote this value of $\tau$ by $\tau^G$. 

\paragraph{Designing an Oracle:} Note that an oracle $\mathcal{O}$ is used to calculate the value of $H(\mathcal{S}, \tau)$. We use a sampling based method to approximate this oracle. Specifically, we sample $n_s$ realizations $\tilde{y} (s)$ from the distribution of $y$ and approximate  $H(\mathcal{S}, \tau)$ as $H(\mathcal{S}, \tau) \approx \tau - \frac{1}{n_s\alpha}\sum_{\tilde{y}} [(\tau-f(\mathcal{S},\tilde{y}))_{+}]$. According to \cite[Lemma 4.1]{ohsaka2017portfolio}, if the number of samples is $n_s = O(\frac{1}{\epsilon^2}\log \frac{1}{\delta}), \delta, \epsilon \in (0,1)$, the CVaR approximation error is less than $\epsilon$ with the probability at least $1-\delta$.

\begin{algorithm}[t]
\caption{Sequential Greedy Algorithm for Problem~\ref{pro:cvar_max}.}  
\begin{algorithmic}[1]
\REQUIRE 
\begin{itemize}
\item Ground set $\mathcal{X}$ and matroid  $\mathcal{I}$
\item User-defined risk level $\alpha \in [0, 1]$
\item Range of the parameter $\tau \in [0, \Gamma]$ and discretization stage $\Delta \in (0, \Gamma]$
\item An oracle $\mathcal{O}$ that evaluates $H(\mathcal{S}, \tau)$
\end{itemize}
\ENSURE 
\begin{itemize}
\item Selected set $\mathcal{S}^{G}$ and corresponding parameter $\tau^{G}$
\end{itemize}
~
\STATE $\mathcal{M}\leftarrow\emptyset$ \label{line:initiliaze}
\STATE \textbf{for} $ ~i=\{0,1,\cdots, \ceil{\frac{\Gamma}{\Delta}}\}$ \textbf{do} \label{line:search_tau_forstart}
\STATE ~~~~$\tau_i = i\Delta$\label{line:search_tau_separation}
\STATE ~~~~$\mathcal{S}_{i}^{G}\leftarrow\emptyset$ \label{line:gre_empty}
\STATE~~~~\textbf{while} \label{line:gre_while_start}
$\mathcal{S}_{i}^{G} \in \mathcal{I}$ \textbf{do}
\STATE~~~~~~~~$s = \underset{s\in \mathcal{X}\setminus \mathcal{S}_{i}^{G}, \mathcal{S}_{i}^{G}\cup \{s\} \in \mathcal{I}}{\text{argmax}}~H((\mathcal{S}_{i}^{G}\cup \{s\}), \tau_i) - H(\mathcal{S}_{i}^{G}, \tau_i)$ \label{line:gre_while_margin}
\STATE~~~~~~~~$\mathcal{S}_{i}^{G}\leftarrow \mathcal{S}_{i}^{G} \cup \{s\}$ \label{line:gre_while_pick}
\STATE~~~~\textbf{end while}\label{line:gre_while_end}
\STATE~~~~ $\mathcal{M} = \mathcal{M} \cup \{(\mathcal{S}_{i}^{G}, \tau_i)\}$\label{line:pair_collection}
\STATE \textbf{end for}\label{line:search_tau_forend}
\STATE $(\mathcal{S}^{G}, \tau^{G}) = \underset{(\mathcal{S}_{i}^{G}, \tau_i) \in \mathcal{M}}{\text{argmax}}~{H(\mathcal{S}_{i}^{G}, \tau_i)}$ \label{line:find_best_pair}

\end{algorithmic}
\label{alg:sga}
\end{algorithm}


\subsection{Performance Analysis of SGA}\label{subsec:analysis_alg}

\begin{theorem}
Let $\mathcal{S}^{G}$ and $\tau^{G}$ be the set and the scalar chosen by the \text{SGA}, and let the $\mathcal{S}^{\star}$ and $\tau^{\star}$ be the set and the scalar chosen by the OPT, we have
\begin{eqnarray}
H(\mathcal{S}^{G},{\tau}^{G}) &\geq& \frac{1}{1+k_f}(H(\mathcal{S}^{\star},\tau^{\star}) - \Delta)
-\frac{k_f}{1+k_f}\Gamma(\frac{1}{\alpha} -1),
\label{eqn:appro_bound_theorem}
\end{eqnarray}
\rev{where $k_f \in [0, 1]$ is the curvature of the $H(\mathcal{S}, \tau)$ in set $\mathcal{S}$. Please see the detailed definition of the curvature in the appendix.} The computational time is $O(\ceil{\frac{\Gamma}{\Delta}} |\mathcal{X}|^{2} n_s)$ where $\Gamma$ and $\Delta$ are the upper bound on $\tau$ and searching separation parameter, $|\mathcal{X}|$ is the cardinality of the ground set $\mathcal{X}$ and $n_s$ is the number of the samplings used by the oracle. 
\label{thm:appro_bound_compu}
\end{theorem}
SGA gives $1/(1+k_f)$ approximation of the optimal with two approximation errors. One approximation error comes from the searching separation $\Delta$. We can make this error very small by setting $\Delta$ to be close to zero with the cost of increasing the computational time. The second approximation error comes from the additive term, 
\begin{equation}
H^{\text{add}} = \frac{k_f}{1+k_f}\Gamma(\frac{1}{\alpha} -1),
\label{eqn:add_error}
\end{equation}
which depends on the curvature $k_f$ and the risk level $\alpha$. When the risk level $\alpha$ is very small, this error is very large which means  SGA may not give a good performance guarantee of the optimal. However, if the function $H(\mathcal{S}, \tau)$ is close to modular in $\mathcal{S}$ ($k_f \to 0$), this error is close to zero.  Notably, when $k_f \to 0$ and $\Delta \to 0$, SGA gives a near-optimal solution ($H(\mathcal{S}^{G},{\tau}^{G}) \to H(\mathcal{S}^{\star},\tau^{\star})$).

Next, we prove Theorem~\ref{thm:appro_bound_compu}. 
We start with the proof of approximation ratio, then go to the analysis of the computational time. We first present the necessary lemmas for the proof of the approximation ratio.

\begin{lemma}
Let $\mathcal{S}_{i}^{\star}$ be the optimal set for a specific $\tau_i$ that maximizes $H(\mathcal{S}, \tau=\tau_i) $. By sequentially searching for $\tau \in [0, \Gamma]$ with a separation $\Delta$, we have 
\begin{equation}
\underset{i\in \{0,1,\cdots, \ceil{\frac{\Gamma}{\Delta}}\}}{\text{max}}H(\mathcal{S}_{i}^{\star}, \tau_i) \geq H(\mathcal{S}^{\star}, \tau^{\star}) -\Delta. 
\label{eqn:tau_sstar_delta}
\end{equation}
\label{lem:tau_sstar}
\end{lemma}

Next, we build the relationship between the set selected by the greedy approach, $\mathcal{S}_{i}^{G}$, and the optimal set $\mathcal{S}_{i}^{\star}$ for $\tau_i$. 
\begin{lemma}
Let $\mathcal{S}_{i}^{\star}$ and $\mathcal{S}_{i}^{G}$  be the sets selected by the greedy algorithm and the optimal approach for a fixed $\tau_i$ that maximizes $H(\mathcal{S}, \tau=\tau_i)$. We have
\begin{eqnarray}
H(\mathcal{S}_{i}^{G},{\tau}_i) &\geq&  \frac{1}{1+k_f} H(\mathcal{S}_{i}^{\star},\tau_i) - \frac{k_f}{1+k_f} \Gamma(\frac{1}{\alpha} -1).
\label{eqn:appro_bound}
\end{eqnarray}
where $k_f$ is the curvature of the function $H(\mathcal{S},\tau)$ in $\mathcal{S}$ with a matroid constraint $\mathcal{I}$. $\Gamma$ is the upper bound of parameter $\tau$.
\label{lem:rela_gre_opt_tau}
\end{lemma}

We leverage Lemma~\ref{lem:tau_sstar} and Lemma~\ref{lem:rela_gre_opt_tau} to prove the approximation ratio in Theorem~\ref{thm:appro_bound_compu}.

\section{Simulations}\label{sec:simulation}
We perform numerical simulations to verify the performance of SGA  in resilient mobility-on-demand and robust environment monitoring. Our code is available online.\footnote{\url{https://github.com/raaslab/risk_averse_submodular_selection.git}}

\subsection{Resilient Mobility-on-Demand under Arrival Time Uncertainty}\label{subsec:mod_demand} 
We consider assigning $R = 6$ supply vehicles to $N=4$ demand locations in a 2D environment. The positions of the demand locations and the supply vehicles are randomly generated within a square environment of 10 units side length.  Denote the Euclidean distance between demand location $i\in \{1,..., N\}$ and vehicle position $j\in \{1,...,R\}$ as $d_{ij}$. Based on the distribution discussion of the arrival efficiency distribution in Section~\ref{subsubsec:environment_monitoring}, we assume each arrival efficiency $e_{ij}$ has a uniform distribution with its mean proportional to the reciprocal of the distance between demand $i$ and vehicle $j$. Furthermore, the uncertainty is higher if the mean efficiency is higher. \rev{Note that, the algorithm can handle other, more complex, distributions for arrival times. We use a uniform distribution for ease of exposition.}
Specifically, denote the mean of $e_{ij}$  as $\bar{e}_{ij}$ and set $\bar{e}_{ij} = 10/d_{ij}$. We model the arrival efficiency distribution to be a uniform distribution as follows: 
$$e_{ij} =[\bar{e}_{ij} - \bar{e}_{ij}^{2.5}/\max\{\bar{e}_{ij}\}, \bar{e}_{ij} + \bar{e}_{ij}^{2.5}/\max\{\bar{e}_{ij}\}], $$ 
where $\max\{\bar{e}_{ij}\} = \max_{i,j}e_{ij}, i\in \{1,..., N\}, j\in \{1,...,R\}$. 

From the assignment utility function (Eq.~\ref{eqn:fsy_assign}),  for any realization of $y$, say $\tilde{y}$, $$f(\mathcal{S}, \tilde{y}) : = \sum_{i\in N} \text{max}_{j\in \mathcal{S}_i}\tilde{e}_{ij}$$ where $\tilde{e}_{ij}$ indicates one realization of $e_{ij}$. If all vehicle-demand pairs are independent from each other, $y$ models a multi-independent uniform distribution. We sample $n_s$ times from underlying multi-independent uniform distribution of $y$ and approximate the auxiliary function $H(\mathcal{S}, \tau)$ 
as $$H(\mathcal{S}, \tau) \approx \tau - \frac{1}{n_s\alpha}\sum_{\tilde{y}} [(\tau-\sum_{i\in N} \text{max}_{j\in \mathcal{S}_i}\tilde{e}_{ij})_{+}].$$ We set the upper bound of the parameter $\tau$ as $\Gamma = N \max (\tilde{e}_{ij}), i = \{1,...N\}, j= \{1,...,R\}$, to make sure $\Gamma - f(\mathcal{S},y) \geq 0$. We set the searching separation for $\tau$ as $\Delta =1$.

\begin{figure}[htb]
\centering{
\subfigure[The value of $H(\mathcal{S}^{G}, \tau^{G})$ with respect to several risk levels $\alpha$.]{\includegraphics[width=0.48\columnwidth]{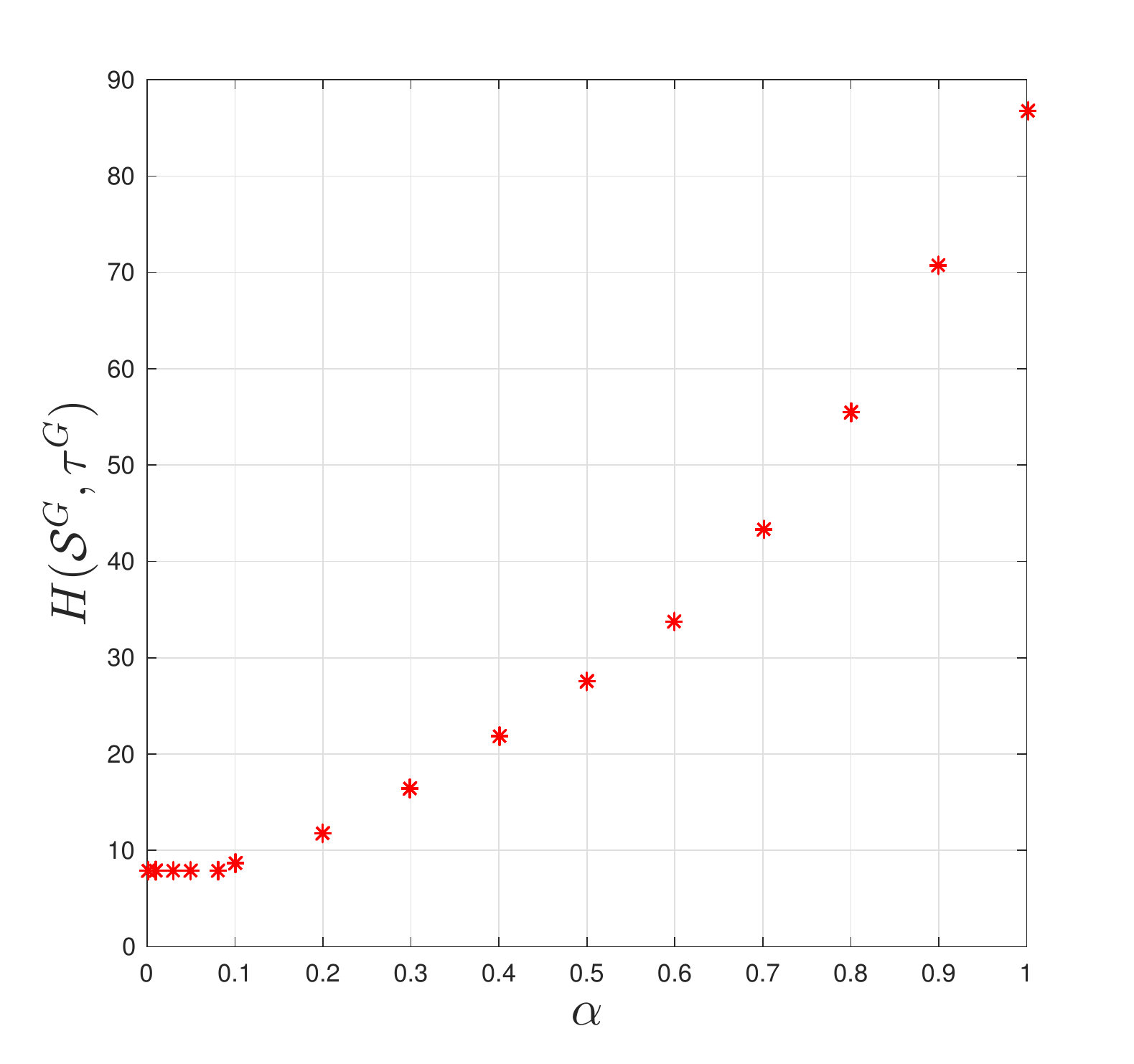}}\goodgap
\subfigure[Function $H(\mathcal{S}^{G}, \tau)$ with respect to several risk levels $\alpha$.]{\includegraphics[width=0.48\columnwidth]{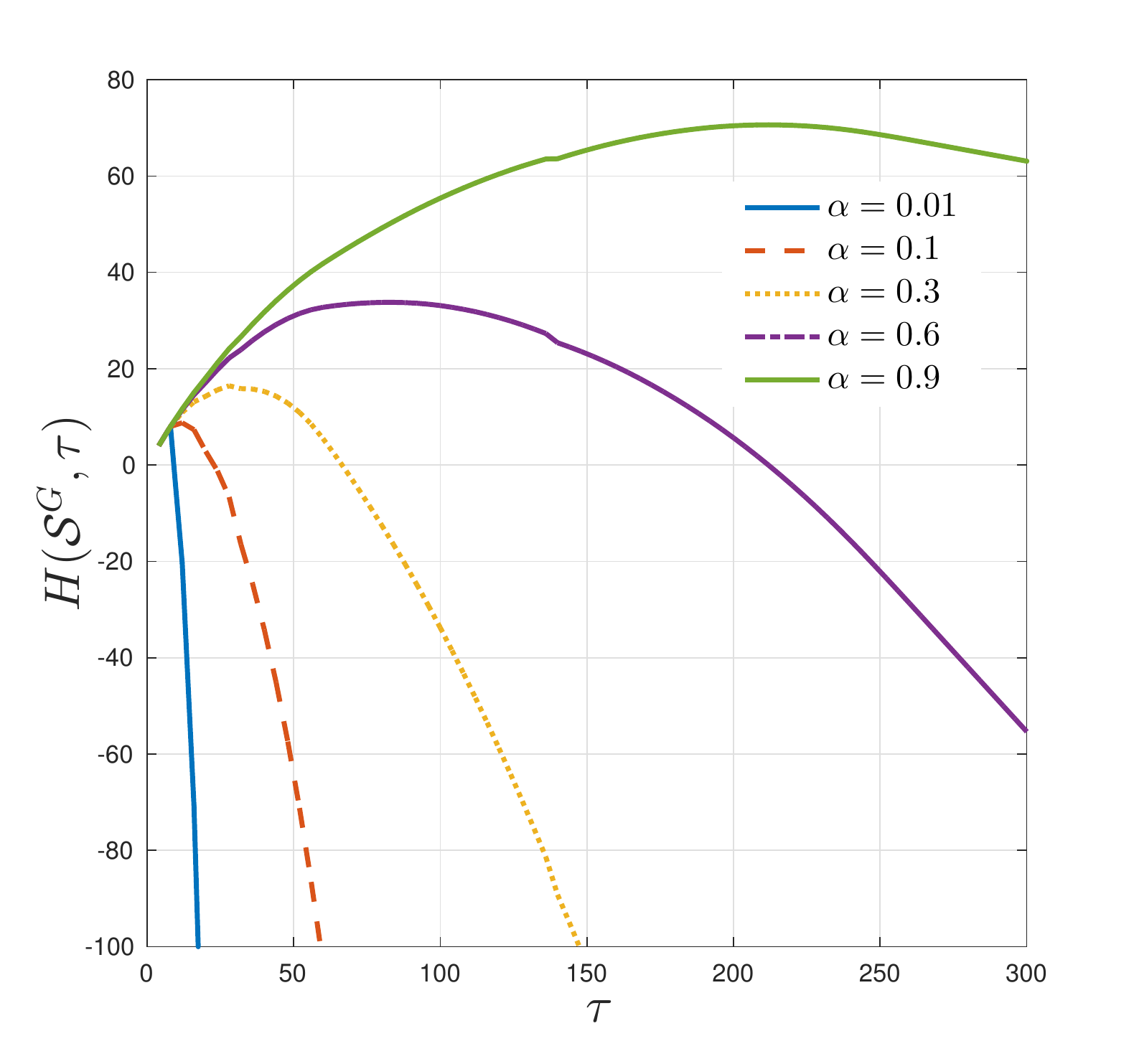}}
\caption{The value of $H(\mathcal{S}, \tau)$ by SGA with respect to different risk confidence levels. 
\label{fig:hsg_alpha_tau}}}
\end{figure}

After receiving the pair $(\mathcal{S}^{G}, \tau^{G})$ from  SGA,  we plot the value of $H(\mathcal{S}^{G}, \tau^{G})$  and  $H(\mathcal{S}^{G}, \tau)$ with respect to different risk levels $\alpha$ in Fig.~\ref{fig:hsg_alpha_tau}. Fig.~\ref{fig:hsg_alpha_tau}-(a) shows that $H(\mathcal{S}^{G}, \tau^{G})$ increases when $\alpha$ increases. This suggests that SGA correctly maximizes $H(\mathcal{S}, \tau)$. Fig.~\ref{fig:hsg_alpha_tau}-(b) shows that $H(\mathcal{S}^{G}, \tau)$ is concave or piecewise concave, which is consistent with the property of $H(\mathcal{S}, \tau)$. 

\begin{figure}
\centering
\begin{minipage}{.48\textwidth}
  \centering
  \includegraphics[width=1\linewidth]{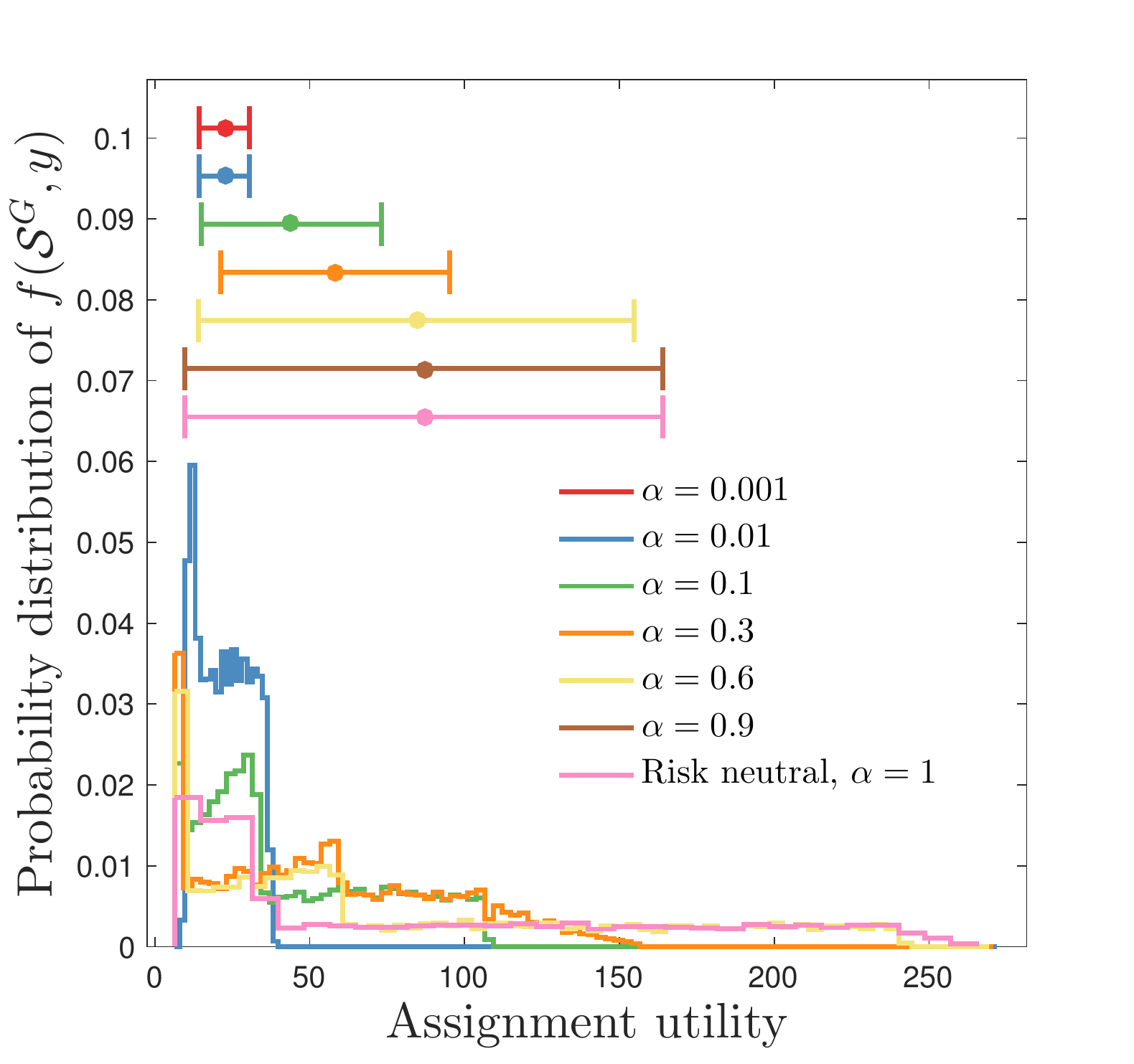}
  \caption{Distribution of the selection utility $f(\mathcal{S}^{G}, y)$ by SGA.}{}
  \label{fig:dis_assign_utility}
\end{minipage} \goodgap
\begin{minipage}{.48\textwidth}
  \centering
  \includegraphics[width=1\linewidth]{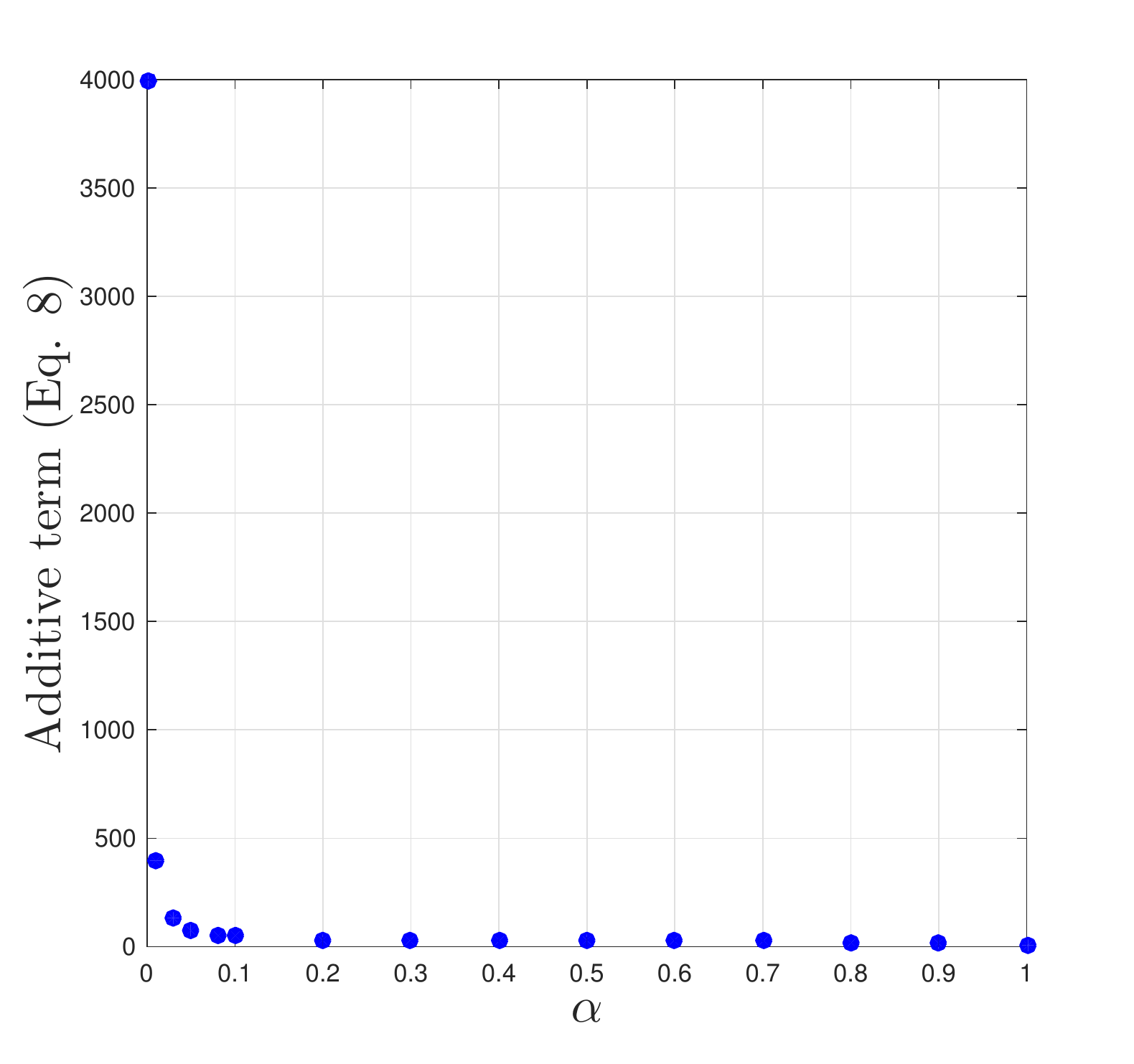}
  \caption{Additive term in the approximation ratio with respect to risk level $\alpha$.}
  \label{fig:add_err_assign}
\end{minipage}
\end{figure}

We plot the distribution of  assignment utility, $$f(\mathcal{S}^{G}, y) =  \sum_{i\in N} \text{max}_{j\in \mathcal{S}_{i}^{G}}e_{ij}$$ in Fig.~\ref{fig:dis_assign_utility} by sampling $n_s=1000$ times from the underlying distribution of  $y$.
$\mathcal{S}_i^{G}$ is a set of vehicles assigned to demand $i$ by SGA. $\mathcal{S}^{G} = \cup_{i=1}^{N} \mathcal{S}_{i}^{G}$. When the risk  level $\alpha$ is small, vehicle-demand pairs with low efficiencies (equivalently, low uncertainties) are selected. \rev{This is because} small risk level indicates the assignment is conservative and only willing to take a little risk. \rev{Thus,} lower efficiency with lower uncertainty is assigned to avoid the risk induced by the uncertainty. In contrast, when $\alpha$ is large, the assignment is allowed to take more risk to gain more assignment utility. Vehicle-demand pairs with high efficiencies (equivalently, high uncertainties) are selected in such a case. Note that, when the risk level is close to zero, SGA may not give a correct solution because of a large approximation error (Fig.~\ref{fig:add_err_assign}). However, this error decreases quickly to zero when the risk level increases.

\begin{figure}[htb]
\centering{
\subfigure[Assignment when $\alpha = 0.1$.]{\includegraphics[width=0.31\columnwidth]{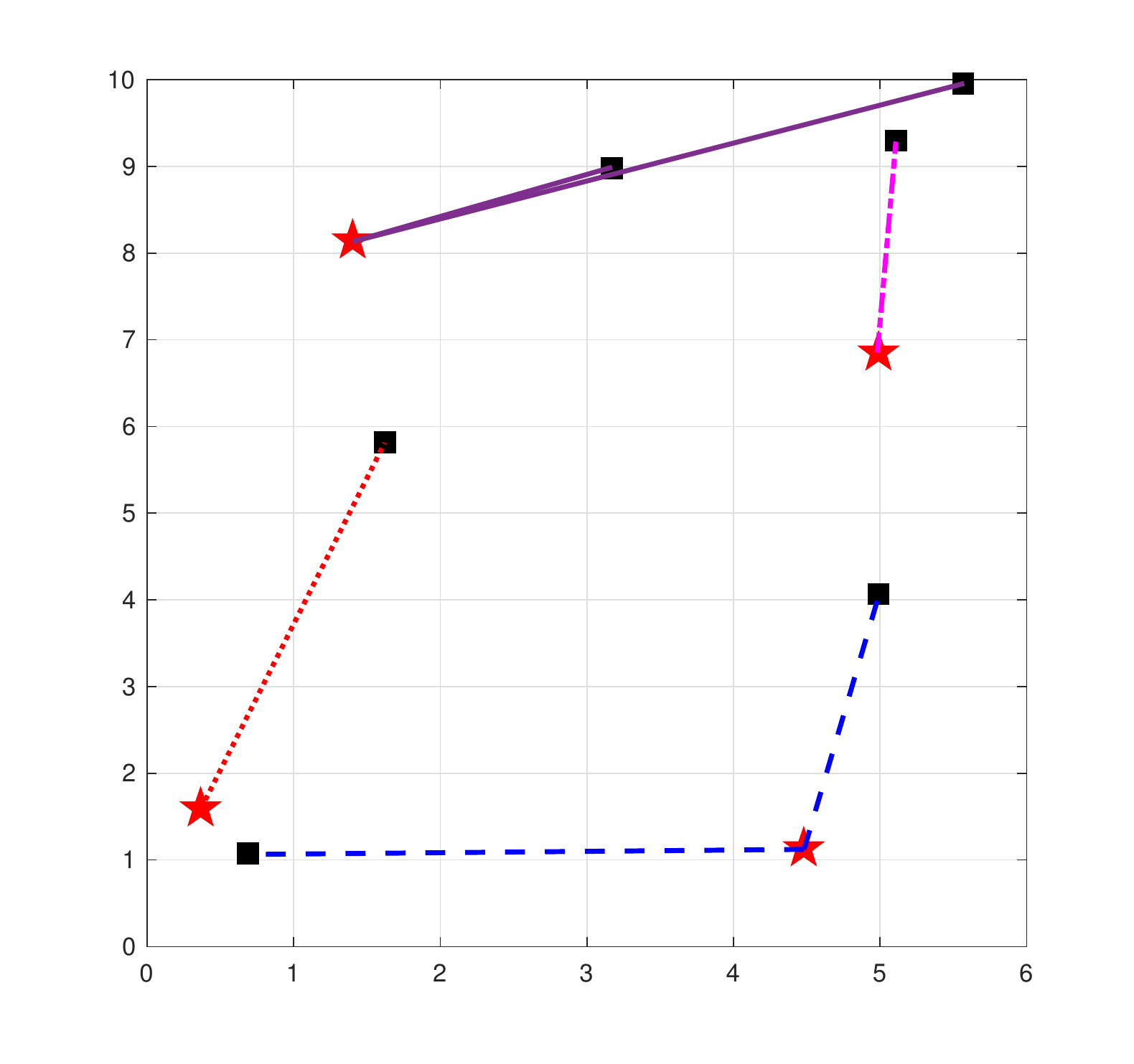}}\goodgap
\subfigure[Assignment when $\alpha = 1$ ($\text{Risk-neutral})$.]{\includegraphics[width=0.31\columnwidth]{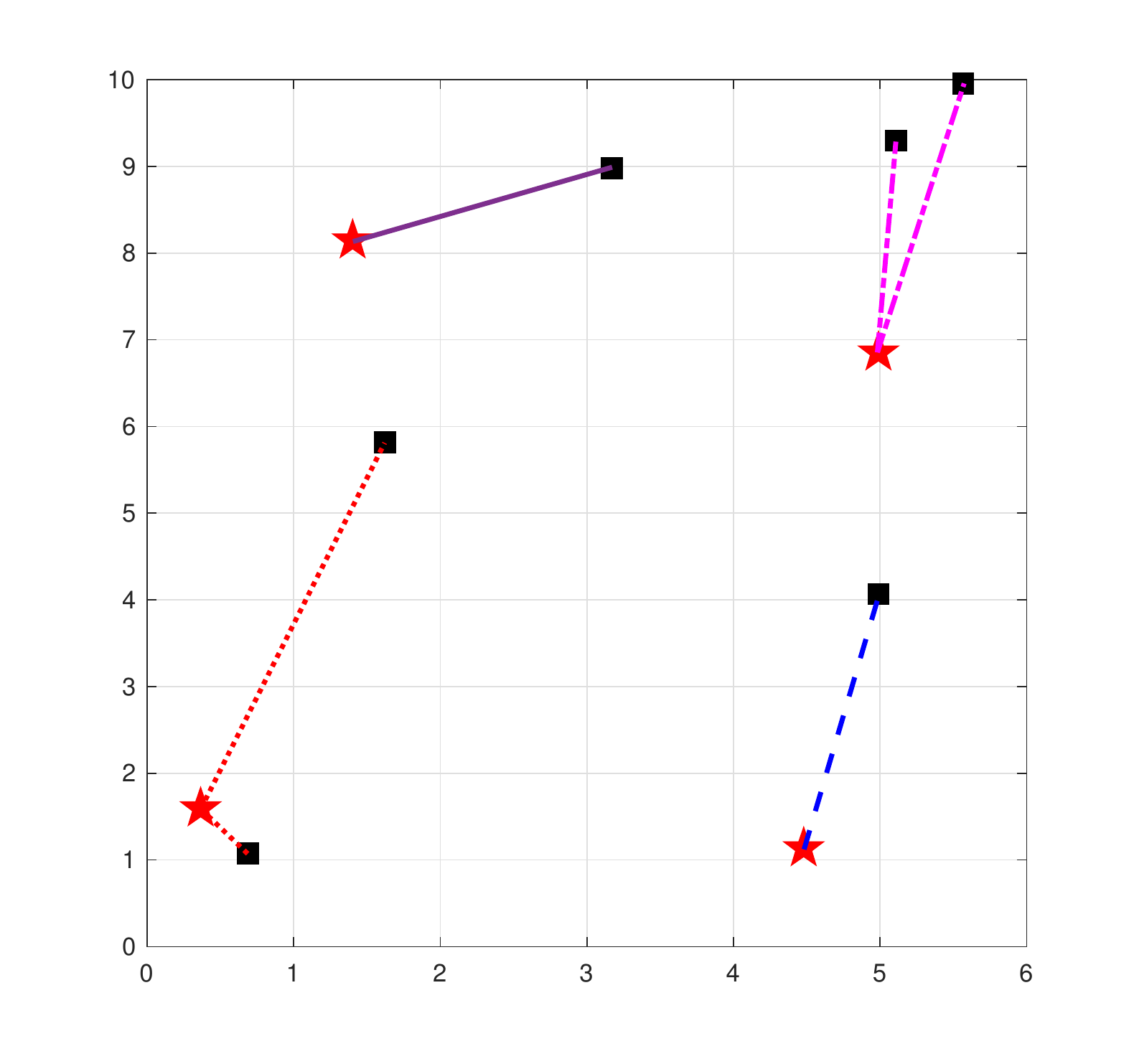}}\goodgap
\subfigure[Assignment utility distributions at $\alpha=0.1$ and $\alpha =1$.]{\includegraphics[width=0.31\columnwidth]{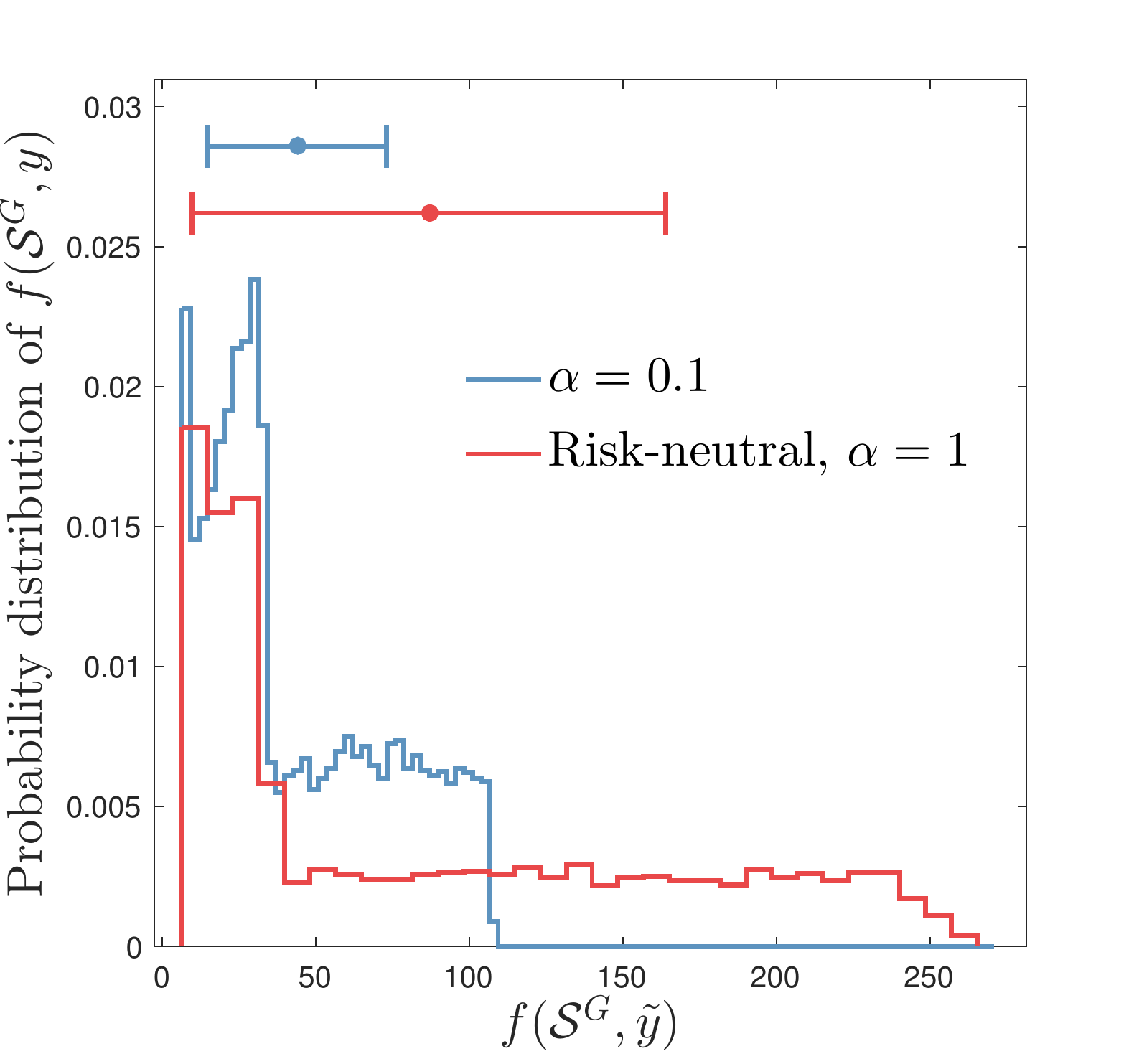}}
\caption{Assignments and utility distributions by  SGA with two extreme risk level values. The red solid \rev{star} represents the demand location. The black solid square represents the vehicle position. The line between the vehicle and the demand represents an assignment. 
\label{fig:two_exteme_assign}}}
\end{figure}

We also compare SGA with CVaR measure with the greedy algorithm with the expectation, i.e., risk-neutral measure~\cite{prorok2018supermodular} in Fig.~\ref{fig:two_exteme_assign}. Note that risk-neutral measure is a special case of $\text{CVaR}_{\alpha}(\mathcal{S})$ measure when $\alpha =1$. We give an illustrative example of the assignment by SGA for two extreme risk levels, $\alpha = 0.1$ and $\alpha = 1$. When $\alpha$ is small ($\alpha = 0.1$), the assignment is  conservative and thus further vehicles (with lower efficiency and lower uncertainty) are assigned to each demand (Fig.~\ref{fig:two_exteme_assign}-(a)). In contrast, when $\alpha =1$, nearby vehicles (with higher efficiency and higher uncertainty) are selected for the demands (Fig.~\ref{fig:two_exteme_assign}-(b)). Even though the mean value of the assignment utility distribution is larger at $\alpha =1$ than $\alpha =0.1$, it is exposed to the risk of receiving lower utility since the mean-std bar at $\alpha =1$ has smaller left endpoint than the mean-std bar at $\alpha =0.1$ (Fig.~\ref{fig:two_exteme_assign}-(c)).

\subsection{Robust Environment Monitoring}\label{subsec:sim_envirnment}
We consider selecting $M=4$ locations from $N=8$ candidate locations to place sensors in the environment (Fig.~\ref{fig:environment_monitor}). Denote the area of the free space as $v^{\text{free}}$. The positions of $N$ candidate locations are randomly generated within the free space $v^{\text{free}}$. We calculate the visibility region for each sensor $v_i$ by using the VisiLibity library~\cite{VisiLibity1:2008}.  Based on the sensor model discussed in Section~\ref{subsec:case_study}, we set the probability of success for each sensor $i$ as 
$$p_i = 1 - v_i/v^{\text{free}},$$
and model the working of each sensor as a Bernoulli distribution with $p_i$ probability of success and $1-p_i$ probability of failure. Thus the random polygon monitored by each sensor $A_i$, follows the distribution 
\begin{equation}
    \begin{cases}
      \text{Pr}[A_i = v_i] = p_i,\\      
      \text{Pr}[A_i = 0] = 1- p_i.
    \end{cases} 
    \label{eqn:cover_dis}
\end{equation}

From the assignment utility function (Eq.~\ref{eqn:fsy_covering}),  for any realization of $y$, say $\tilde{y}$, $$f(\mathcal{S}, y)  = \text{area}(\bigcup_{i=1:M} \tilde{A}_i),$$ where $\tilde{A}_{i}$ indicates one realization of $A_{i}$ by sampling $y$. If all sensors are independent of each other, we can model the working of a set of sensors as a multi-independent Bernoulli distribution. We sample $n_s =1000$ times from the underlying multi-independent Bernoulli distribution of $y$ and  approximate the auxiliary function $H(\mathcal{S}, \tau)$ 
as $$H(\mathcal{S}, \tau) \approx \tau - \frac{1}{n_s\alpha}\sum_{\tilde{y}} [(\tau-\bigcup_{i=1:M} \tilde{A}_i)_{+}],$$ where $\tilde{A}_i$ is one realization of $A_i$ by sampling $y$. We set the upper bound for the parameter $\tau$ as the area of all the free space $v^{\text{free}}$ and set the searching separation for $\tau$ as $\Delta =1$. 
\begin{figure}[h!]
\centering{
\subfigure[The value of $H(\mathcal{S}^{G}, \tau^{G})$ with respect to different risk levels $\alpha$.]{\includegraphics[width=0.48\columnwidth]{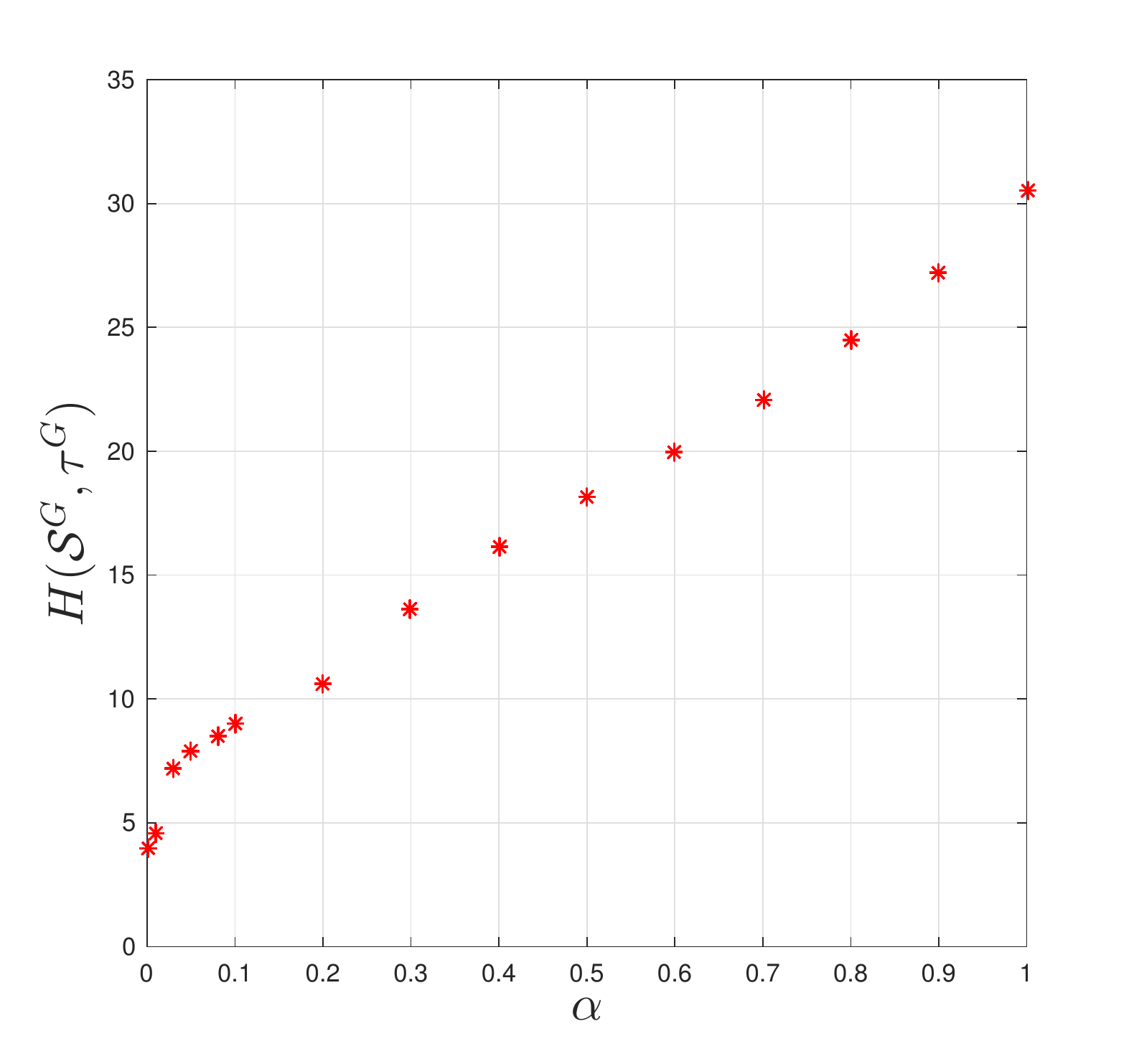}}\goodgap
\subfigure[Function $H(\mathcal{S}^{G}, \tau)$ with respect to several risk levels $\alpha$.]{\includegraphics[width=0.48\columnwidth]{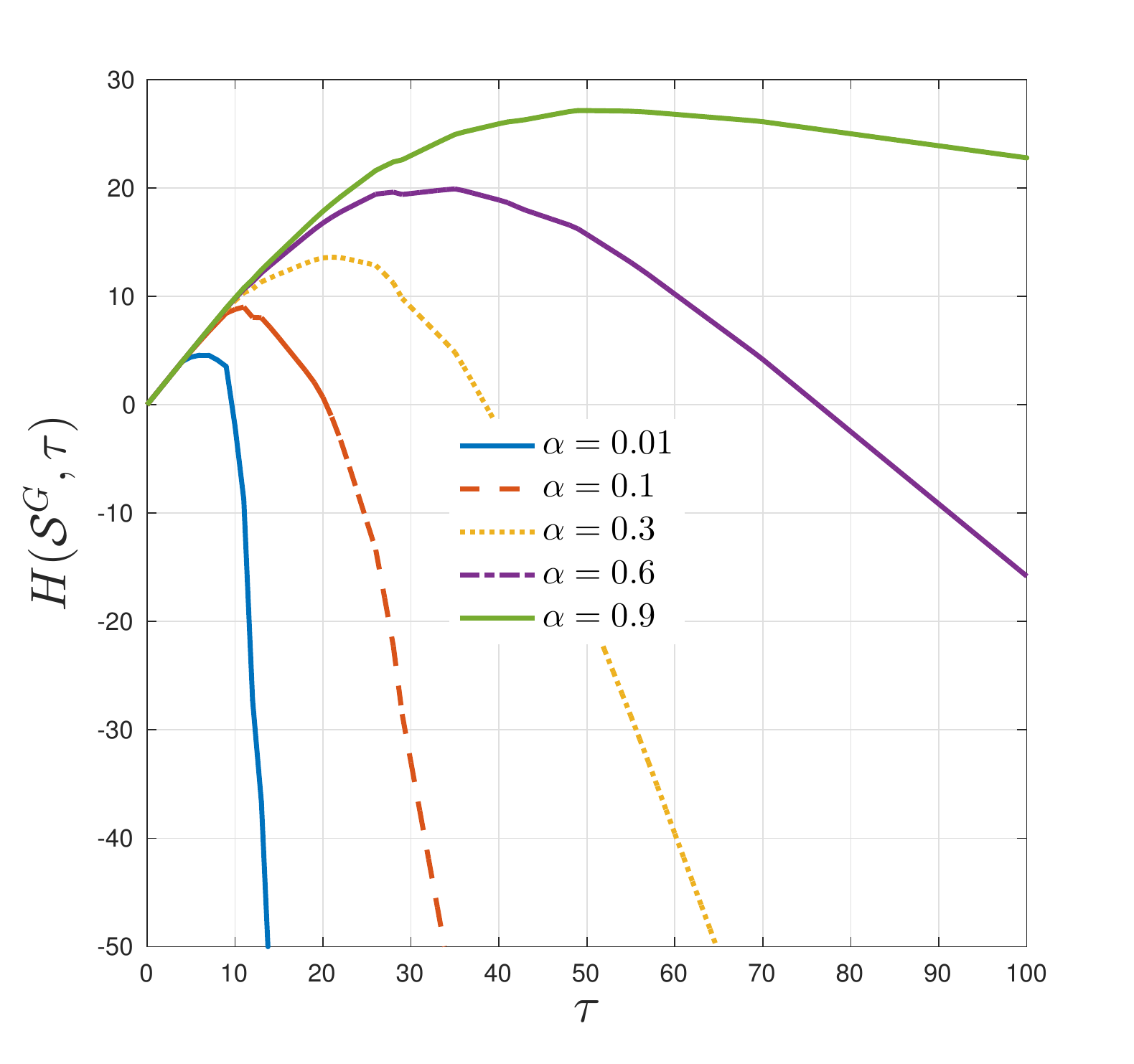}}
\caption{The value of $H(\mathcal{S}, \tau)$ by SGA with respect to different risk confidence levels. 
\label{fig:hsg_alpha_tau_vis}}}
\end{figure}

We use SGA to find the pair $(\mathcal{S}^{G}, \tau^{G})$ with respect to several risk levels $\alpha$. We plot the value of $H(\mathcal{S}^{G}, \tau^{G})$ for several risk levels in Fig.~\ref{fig:hsg_alpha_tau_vis}-(a).  A larger risk level gives a larger $H(\mathcal{S}^{G}, \tau^{G})$, which means the pair $(\mathcal{S}^{G}, \tau^{G})$ found by SGA correctly maximizes
$H(\mathcal{S}, \tau)$ with respect to the risk level $\alpha$. Moreover, we plot functions $H(\mathcal{S}^{G}, \tau)$ for several risk levels $\alpha$ in Fig.~\ref{fig:hsg_alpha_tau_vis}-(b). Note that $\mathcal{S}^{G}$ is computed by SGA at each $\tau$. For each $\alpha$, $H(\mathcal{S}^{G}, \tau)$ shows the concavity or piecewise concavity of function $H(\mathcal{S}, \tau)$. 


\begin{figure}[h!]
\centering
\begin{minipage}[t]{.48\textwidth}
  \centering
  \includegraphics[width=1\linewidth]{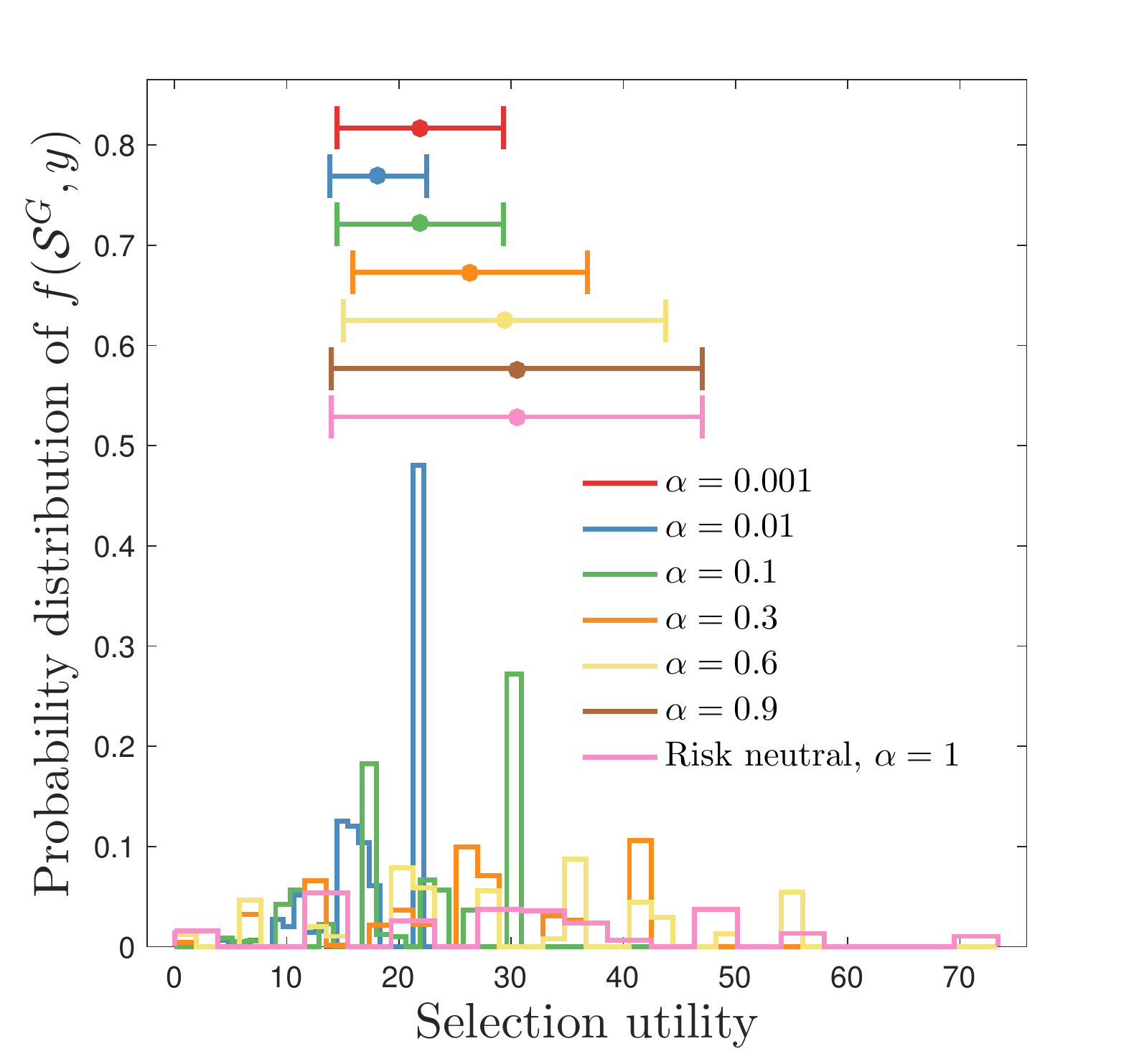}
  \caption{Distribution of the selection utility $f(\mathcal{S}^{G}, y)$ by SGA.}
  \label{fig:dis_select_utility}
\end{minipage}\goodgap
\begin{minipage}[t]{.48\textwidth}
  \centering
  \includegraphics[width=1\linewidth]{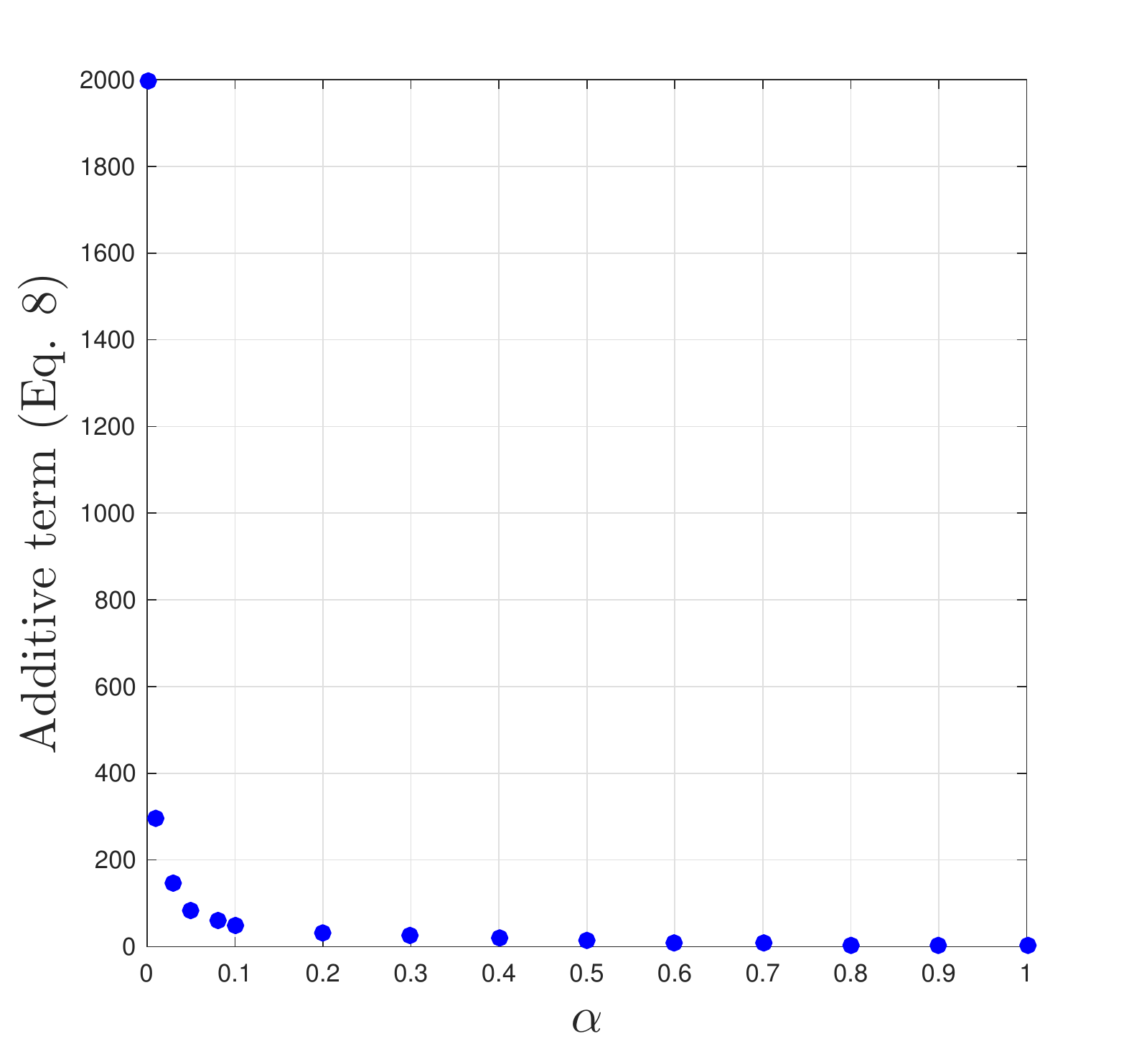}
  \caption{Additive term in the approximation ratio with respect to risk level $\alpha$.}{}
  \label{fig:add_err}
\end{minipage}
\end{figure}

Based on the $\mathcal{S}^{G}$ calculated by  SGA, we sample $n_s = 1000$ times from the underlying distribution of  $y$ and plot the distribution of the selection utility, $$f(\mathcal{S}^{G}, y)  = \bigcup_{i=1:M} A_i, i\in\mathcal{S}^{G}$$ in Fig.~\ref{fig:dis_select_utility}. Note that, when the risk level $\alpha$ is small, the sensors with smaller visibility region and a higher probability of success should be selected. Lower risk level suggests a  conservative selection.  Sensors with a higher probability of success are selected to avoid the risk induced by  sensor failure. In contrast, when $\alpha$ is large, the selection would like to take more risk to gain more monitoring utility. The sensors with larger visibility region and a lower probability of success should be selected.  Fig.~\ref{fig:dis_select_utility} demonstrates this behavior except between
 $\alpha = 0.001$ to $\alpha = 0.01$. This is because when $\alpha$ is very small, the approximation error (Eq.~\ref{eqn:add_error}) is very large as shown in Fig.~\ref{fig:add_err}, and thus SGA may not give a good solution. 

\begin{figure}[htb]
\centering{
\subfigure[Selection when $\alpha = 0.1$.]{\includegraphics[width=0.3\columnwidth]{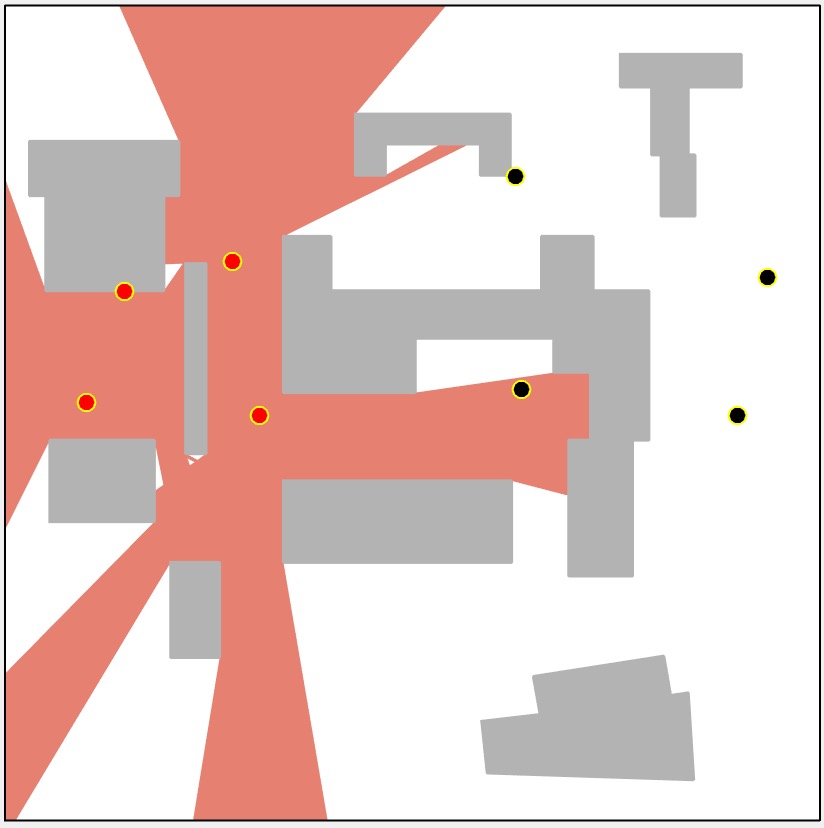}} \goodgap
\subfigure[Selection when $\alpha = 1$ ($\text{Risk-neutral}$).]{\includegraphics[width=0.3\columnwidth]{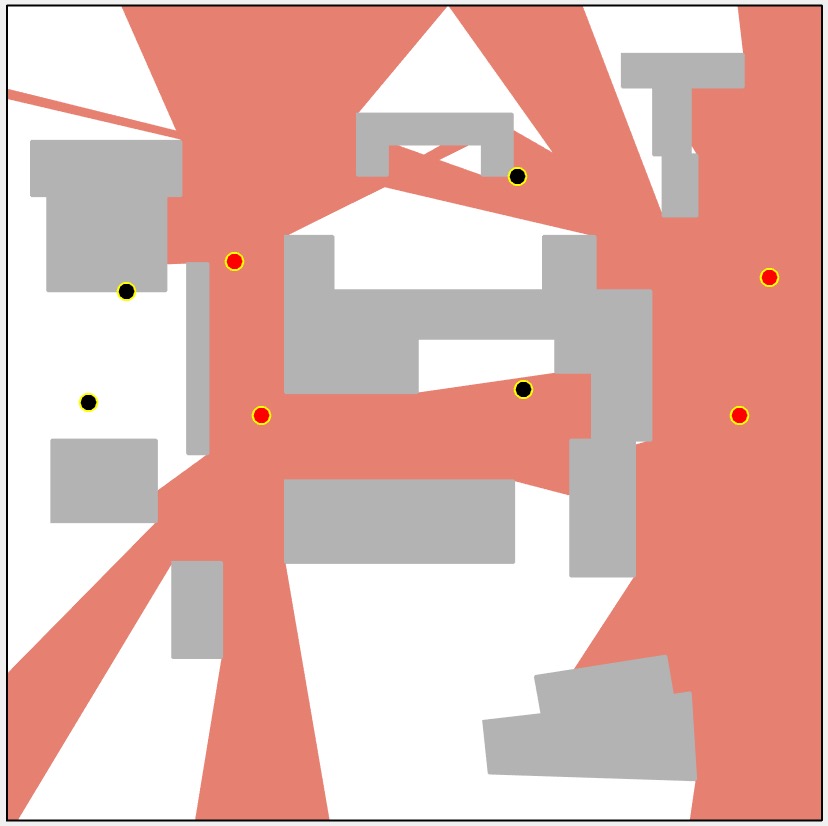}}\goodgap
\subfigure[Selection utility distributions at $\alpha=0.1$ and $\alpha =1$.]{\includegraphics[width=0.32\columnwidth]{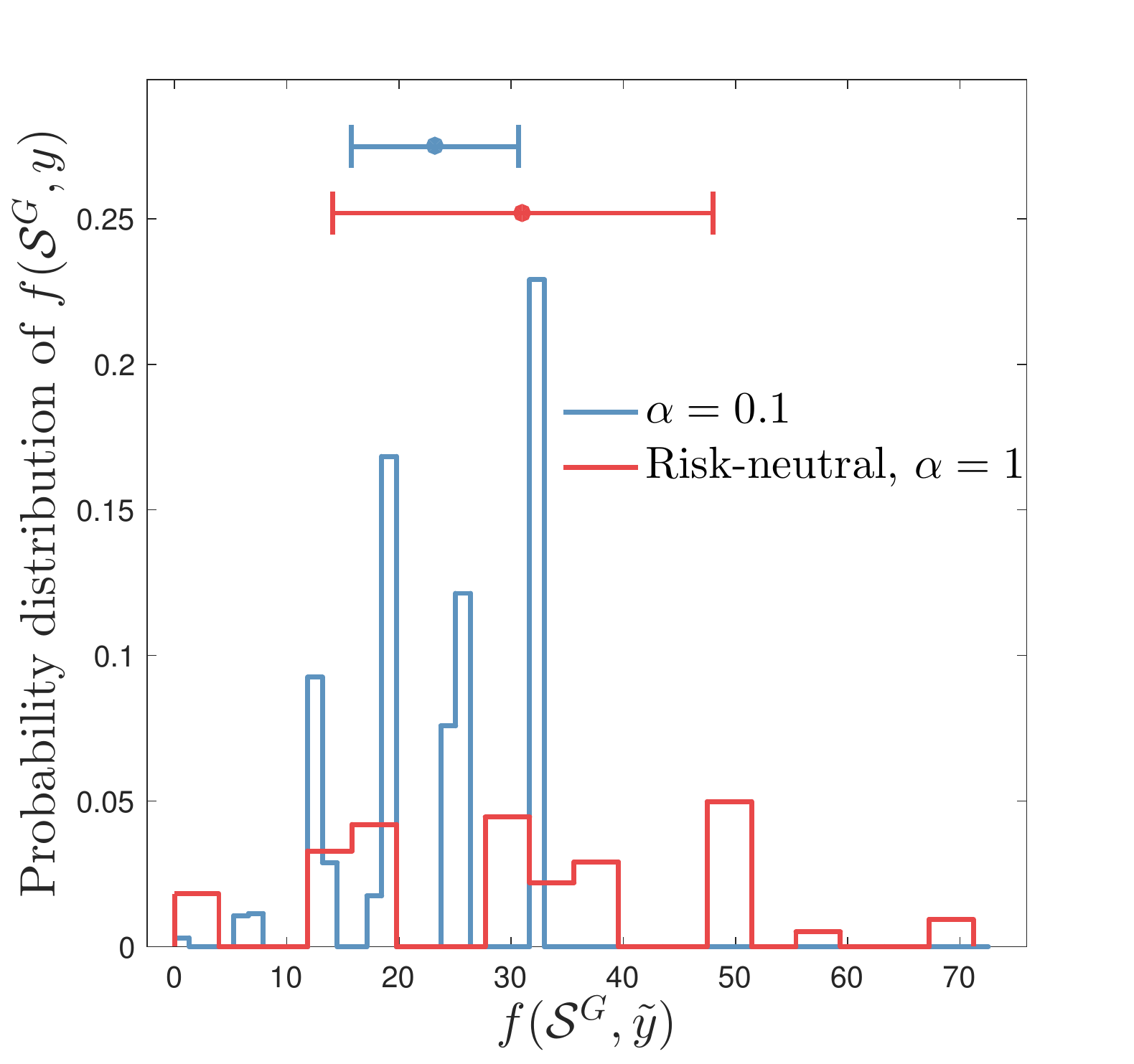}}
\caption{Sensor selection and utility distributions by  SGA with two extreme risk level values. The red solid circle represents the sensor selected by  SGA. 
\label{fig:two_exteme_visi}}}
\end{figure}

We also compare SGA by using CVaR measure with the greedy algorithm by using the expectation, i.e., risk-neutral measure (mentioned in~\cite[Section 6.1]{krause2008near}) in Fig.~\ref{fig:two_exteme_visi}. In fact, the risk-neutral measure is equivalent to case of $\text{CVaR}_{\alpha}(\mathcal{S})$ when $\alpha =1$. We give an illustrative example of the sensor selection by SGA for two extreme risk levels, $\alpha = 0.1$ and $\alpha = 1$. When risk level $\alpha$ is small ($\alpha = 0.1$), the selection is conservative and thus the sensors with small visibility region are selected (Fig.~\ref{fig:two_exteme_visi} -(a)). In contrast, when $\alpha =1$, the risk is neutral and the selection is more adventurous, and thus sensors with large visibility region are selected (Fig.~\ref{fig:two_exteme_visi} -(b)). The mean-std bars of the selection utility distributions in Fig.~\ref{fig:two_exteme_visi} -(c) show that the selection utility at the expectation ($\alpha =1$) has larger mean value than the selection at $\alpha =0.1$. However, the selection at $\alpha =1$ has the risk of gaining lower utility since the left endpoint of mean-std bar at $\alpha =1$ is smaller than the left endpoint of mean-std bar at $\alpha =0.1$.


%
%

\section{Conclusion and Discussion}\label{sec:conclue}

We studied a risk-averse discrete submodular maximization problem. We provide the first positive results for discrete CVaR submodular maximization for selecting a set under matroidal constraints. In particular, we proposed the Sequential Greedy Algorithm and analyzed its approximation ratio and the running time. We demonstrated the two practical use-cases of the CVaR submodular maximization problem. 

Notably, our Sequential Greedy Algorithm works for any matroid constraint. In particular, the multiplicative approximation ratio can be improved to $1/k_f(1-e^{-k_f})$ if we know that the constraint is a uniform matroid~\cite[Theorem 5.4]{conforti1984submodular}. 

The additive term in our analysis depends on $\alpha$. This term can be large when the risk level $\alpha$ is very small. Our ongoing work is to remove this dependence on $\alpha$, perhaps by designing another algorithm specifically for low risk levels. We note that if we use an optimal algorithm instead of the greedy algorithm as a subroutine, then the additive term disappears from the approximation guarantee. \rev{The algorithm also requires knowing $\Gamma$. We showed how to find $\Gamma$ (or an upper bound for it) for the two case studies considered in this paper. Devising a general strategy for finding $\Gamma$ is part of our ongoing work.}

Our second line of ongoing work focuses on applying the risk-averse strategy to multi-vehicle routing, patrolling, and informative path planning in dangerous environments~\cite{jorgensen2018team} and mobility on demand with real-world data sets (2014 NYC
Taxicab Factbook).\footnote{\url{http://www.nyc.gov/html/tlc/downloads/pdf/2014_taxicab_fact_book.pdf}}

\section{Acknowledgements}\label{sec:Acknow}
This work was supported by NSF award IIS-1637915 and ONR Award N00014-18-1-2829. 

\input{appendix.tex}

\bibliographystyle{unsrt} 
\bibliography{refs/myrefs.bib}

\end{document}

%% file: appendix.tex
\section{Appendix}\label{sec:appendix} 
Background on set functions:
\subsection{Monotonicity, Submodularity,  Matroid and Curvature}
We begin by reviewing useful properties of a set function $f(\mathcal{S})$ defined for a finite ground set $\mathcal{X}$ and matroid constraints.

\vspace{10pt}\noindent\textbf{Monotonicity~{\cite{nemhauser1978analysis}}:}  A set function $f: 2^{\mathcal{X}} \mapsto \mathbb{R}$ is monotone (non-decreasing) if and only if for any sets $\mathcal{S}\subseteq \mathcal{S}'\subseteq \mathcal{X}$, we have $f(\mathcal{S})\leq f(\mathcal{S}')$.

\vspace{10pt}\noindent\textbf{Normalized Function~\cite{fisher1978analysis}:}
 A set function $f: 2^{\mathcal{X}} \mapsto \mathbb{R}$ is called normalized if and only if  $f(\emptyset) = 0$.

\vspace{10pt}\noindent\textbf{Submodularity~{\cite[Proposition 2.1]{nemhauser1978analysis}}:} A set function $f: 2^{\mathcal{X}} \mapsto \mathbb{R}$ is submodular if and only if for any sets $\mathcal{S}\subseteq \mathcal{S}'\subseteq \mathcal{X}$, and any element $s\in \mathcal{X}$ and $s \notin \mathcal{S}' $, we have: $f(\mathcal{S}\cup \{s\})- f(\mathcal{S}) \geq f(\mathcal{S}'\cup \{s\})- f(\mathcal{S}')$. Therefore the marginal gain $f(\mathcal{S}\cup \{s\})- f(\mathcal{S})$ is non-increasing. 

\vspace{10pt}\noindent\textbf{Matroid~{\cite[Section 39.1]{schrijver2003combinatorial}}---}  Denote a non-empty collection of subsets of $\mathcal{X}$ as $\mathcal{I}$. The pair $(\mathcal{X}, \mathcal{I})$ is called a matroid if and only if the following conditions are satisfied:\begin{itemize}
\item for any set $\mathcal{S}\subseteq \mathcal{X}$ it must hold that $\mathcal{S}\in\mathcal{I}$, and for any set $\mathcal{P}\subseteq \mathcal{S}$ it must hold that $\mathcal{P}\in\mathcal{I}$. 
\item for any sets $\mathcal{S}, \mathcal{P}\subseteq \mathcal{X}$ and $|\mathcal{P}| \leq |\mathcal{S}|$, it must hold that there exists an element $s\in \mathcal{S} \backslash\mathcal{P}$ such that $\mathcal{P} \cup \{s\} \in \mathcal{I}$. 
\end{itemize} 
We will use two specific forms of matroids that are reviewed next. 

\vspace{10pt}\noindent\textbf{Uniform Matroid:} A \emph{uniform matroid} is a matroid $(\mathcal{X}, \mathcal{I})$ such that  for a positive integer $\kappa$, $\{\mathcal{S}: \mathcal{S}\subseteq \mathcal{X}, |\mathcal{S}|\leq \kappa\}$. Thus, the uniform matroid only constrains the cardinality of the feasible sets in $\mathcal{I}$. 

\vspace{10pt}\noindent\textbf{Partition Matroid:} A \emph{partition matroid} is a matroid $(\mathcal{X}, \mathcal{I})$ such that for a positive integer $n$, disjoint sets $\mathcal{X}_1, ..., \mathcal{X}_n$ and positive integers $\kappa_1,..., \kappa_n$, $\mathcal{X}\equiv \mathcal{X}_1 \cup \cdots \mathcal{X}_n$ and $\mathcal{I} = \{\mathcal{S}: \mathcal{S} \subseteq \mathcal{X}, |\mathcal{S}\cap\mathcal{X}_i|\leq \kappa_i$ for all $i=1,...,n$\}.   

\vspace{10pt}\noindent\textbf{Curvature~\cite{conforti1984submodular}:} 
consider a matroid $\mathcal{I}$ for $\mathcal{X}$, and a non-decreasing submodular set function $f:2^{\mathcal{X}}\mapsto\mathbb{R}$ such that (without loss of generality) for any element $s \in \mathcal{X}$, $f(s)\neq 0$.  The curvature measures how far $f$ is from submodularity or linearity.  Define \emph{curvature} of $f$ over the matroid $I$ as: \begin{equation}\label{eqn:curvature}
k_f \triangleq 1-\underset{s\in\mathcal{S}, \mathcal{S}\in \mathcal{I}}{\min} \frac{f(\mathcal{S})-f(\mathcal{S}\setminus \{ s\})}{f(s)}.
\end{equation}
Note that the definition of curvature $k_f$ (Equation~\ref{eqn:curvature}) implies that  $0 \leq k_f\leq 1$. Specifically, if $k_f = 0$, it means for all the feasible sets $\mathcal{S} \in \mathcal{X}$, $f(\mathcal{S}) = \sum_{s\in \mathcal{S}} f(s)$. In this case, $f$ is a modular function. In contrast, if $k_f = 1$, then there exist a feasible $\mathcal{S}\in\mathcal{I}$ and an element $s \in \mathcal{X}$ such that $f(\mathcal{S}) = f(\mathcal{S} \setminus \{s\})$. In this case, the element $s$ is redundant for the contribution of the value of $f$ given the set $\mathcal{S}\setminus \{s\}$. 

\subsection{Greedy Approximation Algorithm}
In order to maximize a set function $f$, the greedy algorithm selects each element $s$ of $\mathcal{S}$ based on the maximum marginal gain at each round.

We consider maximizing a normalized monotone submodular set function $f$. For any matroid, the greedy algorithm gives a $1/2$ approximation~\cite{fisher1978analysis}. 
In particular, the greedy algorithm can give a $(1-1/e)$--approximation of the optimal solution under the uniform matroid~\cite{nemhauser1978analysis}. 
If we know the curvature of the set function $f$, we have a $1/(1+k_f)$ approximation for any matroid constraint~\cite[Theorem 2.3]{conforti1984submodular}. That is, 
$$\frac{f(\mathcal{S}^{G})}{f^{\star}} \geq \frac{1}{1+k_f}. $$
where $\mathcal{S^{G}} \in \mathcal{I}$ is the set selected by the greedy algorithm, $\mathcal{I}$ is the uniform matroid and $f^{\star}$ is the function value with optimal solution. 
Note that, if $k_f = 0$, which means $f$ is modular, then the greedy algorithm reaches the optimal. If $k_f = 1$, then we have the $1/2$--approximation.

Proof of Lemma~\ref{lem:auxiliary_function}:
\begin{proof}
$H(\mathcal{S},\tau) = \tau - \frac{1}{\alpha}\mathbb{E}[\text{max}(\tau-f(\mathcal{S},y), 0)]$. Since $f(\mathcal{S},y)$ is monotone increasing and submodular in $\mathcal{S}$, $\text{max}\{\tau-f(\mathcal{S},y), 0\}$ is monotone decreasing and supermodular in $\mathcal{S}$, and its expectation is also monotone decreasing and supermodular in $\mathcal{S}$. Then $H(\mathcal{S},\tau)$ is monotone increasing and submodular in $\mathcal{S}$. 

$H(\emptyset,\tau) = \tau(1-\frac{1}{\alpha})$ given $f(\mathcal{S},y)$ is normalized ($f(\emptyset, y) = 0$). Thus, $H(\mathcal{S},\tau)$ is not necessarily normalized since $\tau$ is not necessarily zero. See a similar proof in~\cite{rockafellar2000optimization,maehara2015risk}. 
\label{proof:sub_}
\end{proof}
\qed
Proof of Lemma~\ref{lem:auxi_concave}:
\begin{proof}
$H(\mathcal{S},\tau) = \tau - \frac{1}{\alpha}\mathbb{E}[\text{max}(\tau-f(\mathcal{S},y), 0)]$. Since $\text{max}(\tau-f(\mathcal{S},y), 0)$ is convex in $\tau$, its expectation is also convex in $\tau$. Then $- \frac{1}{\alpha}\mathbb{E}[\text{max}(\tau-f(\mathcal{S},y), 0)]$ is concave in $\tau$ and $H(\mathcal{S},\tau)$ is concave in $\tau$. 
\label{proof:aux_concave}
\end{proof}
\qed
Proof of Lemma~\ref{lem: gradient_auxiliary_function}:
\begin{proof}
By using the result in ~\cite[Lemma 1 and Proof of Theorem 1]{rockafellar2000optimization}, we know that $H(\mathcal{S},\tau)$ is concave and continuously differentiable with derivative given by
$$\frac{\partial H(\mathcal{S},\tau)}{\partial \tau} = 1 - \frac{1}{\alpha} (1 - \Phi(f(\mathcal{S}, y)))$$ where $\Phi(f(\mathcal{S}, y))$ is the cumulative distribution function of $f(\mathcal{S}, y)$. Thus, $0 \leq \Phi(f(\mathcal{S}, y))) \leq 1$, which proves the lemma. 
\label{proof:gradient_auxiliary_fun}
\end{proof}
\qed
\begin{figure}[htb]
\centering{
\subfigure[$H(\mathcal{S}, \tau)$ is under the red dotted line.]{\includegraphics[width=0.49\columnwidth]{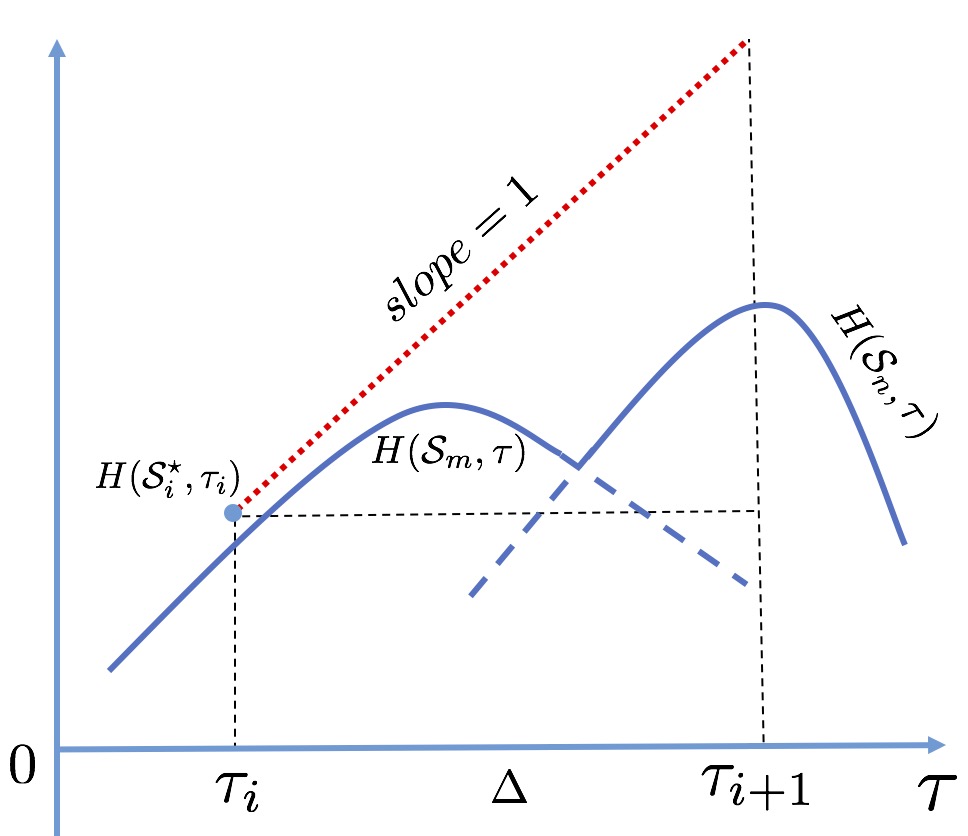}}
\subfigure[$H(\mathcal{S}, \tau)$ is under the red dotted line and green dotted line.]{\includegraphics[width=0.49\columnwidth]{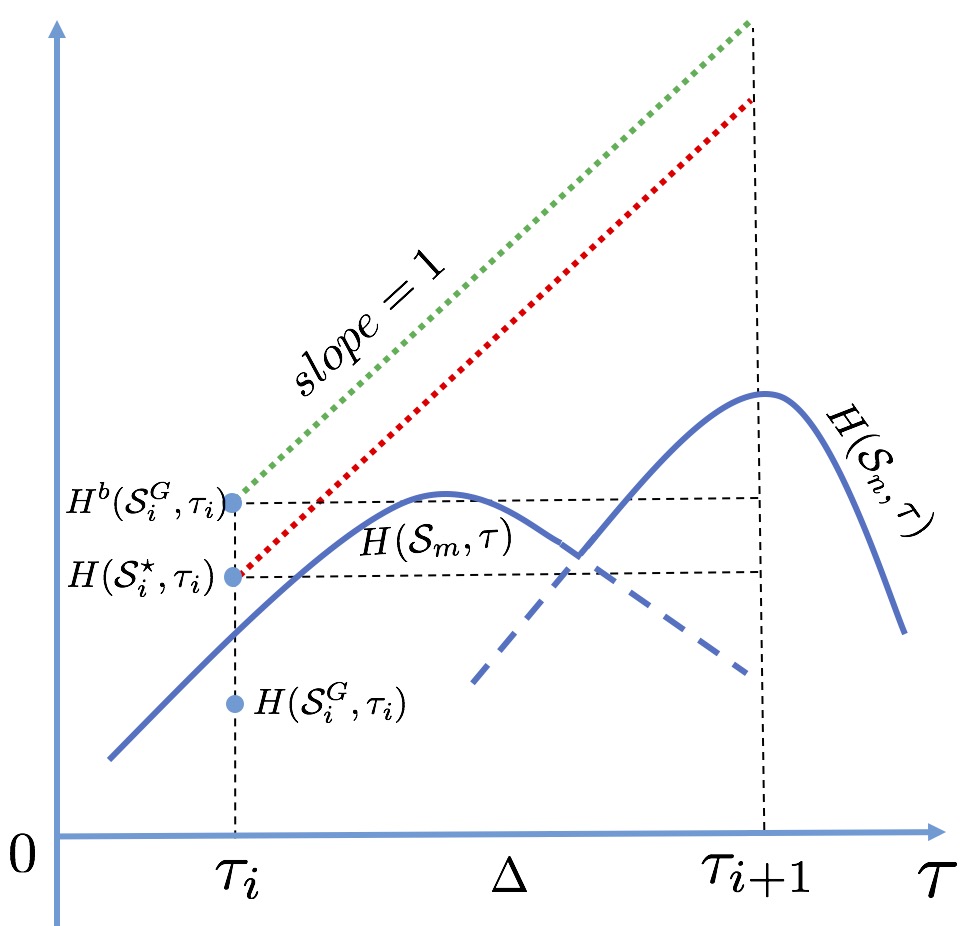}}
\caption{Illustration of $H(\mathcal{S}, \tau)$ within $\tau \in [\tau_i, \tau_{i+1}]$. 
\label{fig:h_tau_ij}}}
\end{figure}
Proof of Lemma~\ref{lem:tau_sstar}:
\begin{proof}
Denote $H_{i}^{\star} = {\text{max}}~H(\mathcal{S}, \tau)$ with ${\tau\in [i\Delta, (i+1)\Delta), \mathcal{S} \in \mathcal{I}}$. From Lemmas~\ref{lem:auxi_concave} and \ref{lem: gradient_auxiliary_function}, we know $H(\mathcal{S},\tau)$ is concave in $\tau$ and $\frac{\partial H(\mathcal{S},\tau)}{\partial \tau} <1$. The properties of concavity and bound on the gradient give 
$$H_{i}^{*} - H(\mathcal{S}_{i}^{\star}, \tau_i) \leq \Delta.$$ We illustrate this claim by using Figure~\ref{fig:h_tau_ij}-(a). Since $\mathcal{S}_{i}^{\star}$ is the optimal set at $\tau_i$ for maximizing $H(\mathcal{S},\tau)$, the value of $H(\mathcal{S},\tau)$ with any other set $\mathcal{S} \in \mathcal{I}$ at $\tau_i$ is at most $H(\mathcal{S}_{i}^{\star}, \tau_i)$. That is, $H(\mathcal{S}, \tau_i)\leq H(\mathcal{S}_{i}^{\star}, \tau_i)$. Since $H(\mathcal{S},\tau)$ is a concave function of $\tau$ for any specific $\mathcal{S}$, $H(\mathcal{S},\tau)$ can be a single concave function, i.e., $H(\mathcal{S}_m, \tau)$ or $H(\mathcal{S}_n, \tau)$ or a piecewise concave function by a combination of several concave functions, i.e., a combination of $H(\mathcal{S}_m, \tau)$ and $H(\mathcal{S}_n, \tau)$ during $\tau \in [\tau_i, \tau_{i+1}]$ (Figure~\ref{fig:h_tau_ij}-(a)). In either case, $H(\mathcal{S}, \tau)$ is below the line starting at $H(\mathcal{S}_{i}^{\star}, \tau_i)$ with $slope = 1$ during $\tau \in [\tau_i, \tau_{i+1}]$ (the red dotted line in Figure~\ref{fig:h_tau_ij}-(a)). Since $H(\mathcal{S}, \tau_i)\leq H(\mathcal{S}_{i}^{\star}, \tau_i)$ and  $H(\mathcal{S},\tau)$ has a bounded gradient $\frac{\partial H(\mathcal{S},\tau)}{\partial \tau} \leq 1$. Thus, $H_{i}^{\star} - H(\mathcal{S}_{i}^{\star}, \tau_i) \leq \frac{\partial H(\mathcal{S},\tau)}{\partial \tau}\Delta = \Delta, ~ \forall i\in \{0,1,\cdots, \ceil{\frac{\Gamma}{\Delta}}\}$. 

Then we have $ H_{i}^{\star} - \text{max}_{i} H(\mathcal{S}_{i}^{\star}, \tau_i)  \leq \Delta, ~ \forall i\in \{0,1,\cdots, \ceil{\frac{\Gamma}{\Delta}}\}$. Note that $H_{i}^{\star}$ is the maximum value of $H(\mathcal{S},\tau)$ at each interval $\tau\in [i\Delta, (i+1)\Delta)$. The maximum value of $H(\mathcal{S},\tau)$, $H(\mathcal{S}^{\star}, \tau^{\star})$ is equal to one of $H_{i}^{\star}, i\in \{0,1,\cdots, \ceil{\frac{\Gamma}{\Delta}}\}$. Thus, we reach the  claim in Lemma~\ref{lem:tau_sstar}.
\label{proof:tau_sstar_delta} 
\end{proof}
\qed
Proof of Lemma~\ref{lem:rela_gre_opt_tau}:
\begin{proof}
We use a the previous result~\cite[Theorem 2.3 ]{conforti1984submodular} for the proof of this claim. We know that for any given $\tau$, $H(\mathcal{S}, \tau)$ is a non-normalized monotone submdoular function in $\mathcal{S}$ (Lemma~\ref{lem:auxiliary_function}). For maximizing normalized monotone submodular set functions, the greedy approach can give a  $1+1/k_f$ approximation of the optimal performance with any matroid constraint~\cite[Theorem 2.3 ]{conforti1984submodular}. After normalizing $H(\mathcal{S}, \tau)$ by $H(\mathcal{S}, \tau) - H(\emptyset, \tau)$, we have 
\begin{equation}
\frac{H(\mathcal{S}_{i}^{G}, \tau_i) - H(\emptyset, \tau_i)}{H(\mathcal{S}_{i}^{\star}, \tau_i) - H(\emptyset, \tau_i)} \geq \frac{1}{1+k_f}, 
\label{eqn:curvature_matroid_constraint}
\end{equation}
with any matroid constraint. Given $0 \leq k_f \leq 1$ and $H(\emptyset, \tau) = -\tau(\frac{1}{\alpha}-1)$, we transform Equation~\ref{eqn:curvature_matroid_constraint} into,
\begin{eqnarray}
H(\mathcal{S}_{i}^{G},{\tau}_i) &\geq&  \frac{1}{1+k_f} H(\mathcal{S}_{i}^{\star},\tau_i) - \frac{k_f}{1+k_f} \tau_i(\frac{1}{\alpha} -1)\nonumber\\
&\geq&\frac{1}{1+k_f} H(\mathcal{S}_{i}^{\star},\tau_i) - \frac{k_f}{1+k_f} \Gamma(\frac{1}{\alpha} -1)
\label{eqn:sgstar_taub}
\end{eqnarray}
where Equation~\ref{eqn:sgstar_taub} holds since $\Gamma$ is the upper bound of $\tau$. Thus, we prove the Lemma~\ref{lem:rela_gre_opt_tau}. 
\label{pro:lem_gre_opt_relative}
\end{proof}\qed
Proof of Theorem~\ref{thm:appro_bound_compu}:
\begin{proof}
From Equation~\ref{eqn:appro_bound} in Lemma~\ref{lem:rela_gre_opt_tau}, we have $H(\mathcal{S}_{i}^{\star},\tau_i)$ is bounded by
\begin{equation}
H(\mathcal{S}_{i}^{\star},\tau_i) \leq (1+k_f)H(\mathcal{S}_{i}^{G},{\tau}_i) + k_f \Gamma(\frac{1}{\alpha} -1). 
\label{eqn:upper_hstar_tau}
\end{equation}

Denote this upper bound as $$H^{b}(\mathcal{S}_{i}^{G},\tau_i) : = (1+k_f)H(\mathcal{S}_{i}^{G},{\tau}_i) + k_f \Gamma(\frac{1}{\alpha} -1).$$
We know $H(\mathcal{S}, \tau)$ is below the line starting at $H(\mathcal{S}_{i}^{\star}, \tau_i)$ with $slope = 1$ during $\tau \in [\tau_i, \tau_{i+1}]$ (the red dotted line in Figure~\ref{fig:h_tau_ij}-(a)/(b)) (Lemma~\ref{lem:tau_sstar}). $H(\mathcal{S}, \tau)$ must be also below the line starting at $H^{b}(\mathcal{S}_{i}^{G},\tau_i)$ with $slope = 1$ during $\tau \in [\tau_i, \tau_{i+1}]$ (the green dotted line in Figure~\ref{fig:h_tau_ij}-(b)). Similar to the proof in Lemma~\ref{lem:tau_sstar}, we have $H_{i}^{\star} - H^{b}(\mathcal{S}_{i}^{G}, \tau_i) \leq \Delta$  and 
\begin{equation}
{\text{max}}_{i\in \{0,1,\cdots, \ceil{\frac{\Gamma}{\Delta}}\}} H^{b}(\mathcal{S}_{i}^{G}, \tau_i) \geq H(\mathcal{S}^{\star}, \tau^{\star}) -\Delta. 
\label{eqn: max_bound_hstar}
\end{equation}
SGA selects the pair $(\mathcal{S}^{G},\tau^{G})$ as the pair  $(\mathcal{S}_{i}^{G},\tau_i)$ with $\text{max}_i ~H(\mathcal{S}_{i}^{G},\tau_i)$. Then by Inequalities~\ref{eqn:upper_hstar_tau} and \ref{eqn: max_bound_hstar}, we have
\begin{eqnarray}
(1+k_f)H(\mathcal{S}^{G},{\tau}^{G}) + k_f \Gamma(\frac{1}{\alpha} -1) \geq H(\mathcal{S}^{\star}, \tau^{\star}) -\Delta.
\label{eqn:approx_final_derive}
\end{eqnarray}
By rearranging the terms, we get the approximation ratio in Theorem~\ref{thm:appro_bound_compu}. 

Next, we give the proof of the computational time of  SGA in Theorem~\ref{thm:appro_bound_compu}. 
We verify the computational time of SGA by following the stages of the pseudo code in SGA.  First, from line~\ref{line:search_tau_forstart} to \ref{line:search_tau_forend}, we use a ``for'' loop for searching $\tau$ which takes $\ceil{\frac{\Gamma}{\Delta}}$ evaluations. Second, within the ``For'' loop, we use the greedy algorithm to solve the subproblem (lines~\ref{line:gre_empty}--\ref{line:gre_while_end}). In order to select a subset $\mathcal{S}$ with size $|\mathcal{S}|$  from a ground set $\mathcal{X}$ with size $|\mathcal{X}|$, the greedy algorithm takes  $|\mathcal{S}|$ rounds (line~\ref{line:gre_while_start}), and calculates the marginal gain of  the remaining elements in $\mathcal{X}$ at each round (line~\ref{line:gre_while_margin}). Thus, the greedy algorithm takes $\sum_{i=1}^{|\mathcal{S}|} |\mathcal{X}|-i$ evaluations. Thus, the greedy algorithm takes $|\mathcal{S}|(\sum_{i=1}^{|\mathcal{S}|} |\mathcal{X}|-i)$ evaluations. Third, by calculating the marginal gain for each element, the oracle $\mathcal{O}$ samples $n_s$ times for computing $H(\mathcal{S})$. Thus, overall, the ``for'' loop containing the greedy algorithm with the oracle sampling takes $\ceil{\frac{\Gamma}{\Delta}}|\mathcal{S}|(\sum_{i=1}^{|\mathcal{S}|} |\mathcal{X}|-i)n_s$ evaluations. Last, finding the best pair from storage set $\mathcal{M}$ (line~\ref{line:find_best_pair} of Alg.~\ref{alg:sga}) takes $O(\ceil{\frac{\Gamma}{\Delta}})$ time. Therefore, the computational complexity for SGA is, 
$$\ceil{\frac{\Gamma}{\Delta}}|\mathcal{S}|(\sum_{i=1}^{|\mathcal{S}|} |\mathcal{X}|-i)n_s + O(\ceil{\frac{\Gamma}{\Delta}}) = O(\ceil{\frac{\Gamma}{\Delta}} |\mathcal{X}|^{2} n_s),$$
given $|\mathcal{S}|\leq |\mathcal{X}|$. 
\label{prf:compu_time}
\end{proof}
\qed